\documentclass[runningheads]{llncs}

\usepackage{eccv}

\usepackage{eccvabbrv}

\usepackage{graphicx}
\usepackage{booktabs}

\usepackage[accsupp]{axessibility}  
\usepackage{tikz}
\usetikzlibrary{positioning,arrows.meta,fit}

\usepackage{hyperref}

\usepackage{orcidlink}

\begin{document}

\title{The Second LoViF 2026 Challenge on Real-World All-in-One Image Restoration: Methods and Results}


\author{Xiang Chen\textsuperscript{$*$} \and Hao Li\textsuperscript{$*$} \and Jiangxin Dong\textsuperscript{$*$} \and Jinshan Pan\textsuperscript{$*$} \and  Xin Li\textsuperscript{$*$} \and Hongbo Ding \and Junpeng Jiang \and Xingyu Qiu \and Yilian Zhong \and Yuxiang Chen \and Shibo Yin \and Zixuan Huang \and Yushun Fang \and Xilei Zhu \and Yahui Wang \and Chen Lu \and Xiaodong Zhou \and Qingyue Cao \and Changwei Gong \and Jingyun Liu \and Xingchen Yi \and Hansen Shi \and Ruiyi Liu \and Jirui Xie \and Tao Liu \and Wenzhuo Ma \and Hongzhen Li \and Yongyong Chen \and Zheng Zhou \and Jingyong Su \and Jie Liu \and Haijin Zeng \and Cheng Li \and Peishuai Zha \and Ziyi Wang \and Jian Tang \and Yan Chen \and Long Bao \and Heng Sun \and Jiyuan Zhang \and Shuai Liu \and Wei Ding \and Chengjun Guo \and Yibin Huang \and Xiaotao Wang \and Dongqing Zou \and Lei Lei \and Xiaofeng Wang \and Xiao Liu \and Yulin Wu \and Yuhan Zhao \and Shurui Peng \and Chao Ren \and Yu-Kai Wang \and Kosuke Shigematsu \and Asuka Shin \and Rong-Lin Jian \and Cheng-Jun Kang \and Jin-Hui Jiang \and Jialin Zhou \and Kuo Yuan \and Songyu Zhang \and E B Benson \and Ashfaq Hussain \and Pruthvikanth AC \and Qirui Chen \and Jinyuan Chen \and Jun Zhang \and Xu Zhang \and Xuhui Cao \and Jiaqi Ma \and Laibin Chang \and Yuchun Miao \and Yichu Xu \and Yuanzhi Yao \and Shi Chen \and Yuning Cui \and Huan Zhang \and Lefei Zhang \and Saeed Ahmad \and Ik Hyun Lee \and Jun Young Park \and Ji Hwan Yoon \and Shangquan Sun \and Behrooz Nobahar-Moghanlou \and Majid Edalatjou \and Karim Shahi-Niyar \and Ruibo Zhang \and Dexiang Hong \and Xinyan Liu \and Shengeng Tang \and Weidong Chen}

\authorrunning{X. Chen et al.}

\institute{}


\maketitle

\renewcommand{\thefootnote}{}

\footnotetext{$^{*}$X. Chen(\textcolor{magenta}{chenxiang@njust.edu.cn}), H. Li(\textcolor{magenta}{haoli@njust.edu.cn}), J. Dong(\textcolor{magenta}{jxdong@njust.edu.cn}), and J. Pan(\textcolor{magenta}{jspan@njust.edu.cn}) are the challenge organizers.}

\footnotetext{$^{*}$X. Li(\textcolor{magenta}{xin.li@ustc.edu.cn}) is the workshop organizer in assisting the organization of this challenge.}

\footnotetext{The remaining authors are participants in the Second LoViF 2026 Challenge on Real-World All-in-One Image Restoration.}

\begin{abstract}
This paper presents a review of the second LoViF Challenge on Real-World All-in-One Image Restoration.
The challenge aims to advance unified image restoration under diverse real-world degradation conditions, including blur, low-light, haze, rain, and snow.
It provides a common benchmark for evaluating the restoration accuracy, robustness, and generalization capability of models across multiple degradation categories within a unified framework.
The competition attracted 158 registered participants, and 20 teams were included in the final ranking after their submitted results were successfully reproduced and verified.
This report provides a comprehensive analysis of the submitted solutions and corresponding results, highlighting recent advances in real-world all-in-one image restoration.
The summarized methods and empirical findings reveal effective design strategies and establish an updated benchmark for future research in real-world low-level vision.

\end{abstract}

\section{Introduction}
\label{sec:intro}
All-in-one image restoration seeks to develop a unified model capable of recovering high-quality images from inputs affected by diverse degradations, including blur, low-light, haze, rain, and snow. In contrast to task-specific approaches that independently design and optimize a dedicated model for each degradation, all-in-one methods learn shared representations across multiple restoration tasks and emphasize generalization, adaptability, and scalability within a common framework~\cite{chen2026foundir,jiang2025survey}.

Most existing all-in-one restoration methods construct their training sets by combining multiple task-specific synthetic datasets~\cite{jin2025smokebench,Airnet,potlapalli2023promptir,zheng2024diffuir}. Although such a strategy facilitates unified training, synthetic degradations often differ substantially from those encountered in real-world imaging. Real degradations are affected by complex camera pipelines, environmental conditions, scene content, and illumination distributions, and may exhibit spatially non-uniform or coupled characteristics. Consequently, the domain gap between synthetic and real-world data remains a major obstacle to the practical deployment of unified restoration models.

Recent studies increasingly focus on collecting large-scale real-world paired data for developing more robust restoration models. Li et al.~\cite{li2024foundir} propose FoundIR, a million-scale dataset designed for training image restoration foundation models. By jointly controlling internal camera parameters and external imaging conditions, FoundIR captures well-aligned low-quality and high-quality image pairs through multiple rounds of acquisition and a dedicated alignment criterion. It therefore provides both a substantially larger number of real-world samples and greater degradation diversity than conventional restoration datasets. In addition, WeatherBench~\cite{guan2025weatherbench} provides a real-world benchmark covering diverse adverse-weather conditions, offering complementary data for evaluating unified weather restoration methods.

Building upon the first edition of the challenge~\cite{chen2026lovif}, we organize the LoViF 2026 Second Challenge on Real-World All-in-One Image Restoration in conjunction with the second LoViF Workshop at ECCV 2026. The challenge continues to investigate unified restoration under five common real-world degradation categories, namely blur, low-light, haze, rain, and snow. It provides a common benchmark for evaluating restoration fidelity, perceptual quality, robustness, and cross-degradation generalization. The benchmark, termed FoundIR-LoViF, is curated from the real-world paired data provided by FoundIR~\cite{li2024foundir} and WeatherBench~\cite{guan2025weatherbench}, enabling participants to develop and compare unified restoration models under a consistent experimental setting.

This challenge is held in conjunction with the second LoViF Workshop at ECCV 2026. The workshop hosts a series of challenges covering real-world all-in-one image restoration, unified removal of raindrops and reflections, ultra-low-bitrate image compression, day and night raindrop removal, and AIGC image compression. Together, these challenges provide a platform for advancing practical low-level vision methods under diverse real-world and generative visual processing scenarios.

\section{The Second LoViF 2026 Real-World All-in-One Image Restoration Challenge}
\label{sec:challenge}
Building upon the first edition, the second LoViF 2026 Challenge on Real-World All-in-One Image Restoration focuses on developing unified restoration models for images captured under diverse real-world degradation conditions.
In contrast to conventional task-specific approaches~\cite{chen2023learning,kong2023efficient,yang2026unirain,guan2026harmonizing,yang2026rethinking,guan2026rethinking,yang2026textual}, which design separate models for individual restoration tasks, this challenge requires a common framework to handle multiple degradation categories.
Such a unified setting evaluates whether a restoration model can effectively exploit shared knowledge across different tasks while maintaining sufficient adaptability to the distinct characteristics of each degradation.

\noindent\textbf{Dataset.}
The benchmark data are provided by FoundIR~\cite{li2024foundir} and WeatherBench~\cite{guan2025weatherbench}, covering five representative categories of real-world image degradation, \ie, \textit{blur, low-light, haze, rain, and snow}.
The dataset contains diverse scenes and degradation characteristics and is divided into training, validation, and test sets under a balanced multi-task setting.
For each degradation category, 4,900 paired low-quality and ground-truth images with a resolution of $512 \times 512$ are provided for training, while 100 degraded images are used for validation and another 100 degraded images are reserved for final testing.
Consequently, the complete training set consists of 24,500 paired low-quality and ground-truth images, and the validation and test sets each contain 500 degraded images.

\noindent\textbf{Evaluation protocol.}
The submitted restoration results are evaluated using PSNR, SSIM, and LPIPS, which jointly measure pixel-level reconstruction fidelity, structural similarity, and perceptual quality.
Following the evaluation protocol adopted in the previous edition~\cite{li2025ntire}, the final ranking is determined by a composite score defined as
\begin{equation}
    \text{Score}= \text{PSNR(Y)} + 10 \times \text{SSIM(Y)} -5\times \text{LPIPS},
\end{equation}
where $\text{PSNR(Y)}$ and $\text{SSIM(Y)}$ are computed on the luminance channel after converting the restored and ground-truth images from the RGB color space to the YCbCr color space.
For LPIPS, the pixel values of the restored and reference images are normalized to the range $[-1,1]$, and perceptual distance is calculated using the AlexNet-based configuration.
A higher composite score indicates better overall restoration performance.

By retaining a balanced dataset and a unified evaluation protocol across five real-world degradation categories, the second challenge provides an updated benchmark for comparing recent all-in-one restoration methods under consistent experimental conditions.
It further promotes the development of restoration models with stronger fidelity, perceptual quality, robustness, and cross-degradation generalization in practical scenarios.

\section{Challenge Results}
\label{challenge_results}

The final results of the LoViF 2026 Second Challenge on Real-World All-in-One Image Restoration are presented in Table~\ref{tab:results_1}.
Among the 20 teams included in the final ranking, the top three teams are Re:Pixel, REDnoteMediaLab, and LucidWorld, led by Hongbo Ding, Junpeng Jiang, and Xiaodong Zhou, respectively.
They achieve final scores of 42.28, 41.81, and 41.40.
The score differences between the first- and second-ranked teams and between the second- and third-ranked teams are only 0.47 and 0.41, respectively, indicating intense competition among the leading solutions.
Meanwhile, the third-ranked team outperforms the fourth-ranked team by 6.59 points, showing that the top three methods form a distinct leading group in the final leaderboard.

Despite the strong performance achieved by the leading teams, real-world all-in-one image restoration remains a challenging problem.
The submitted methods exhibit considerable differences in restoration accuracy and reported computational complexity, reflecting diverse trade-offs between model capacity, inference cost, and restoration quality.
In particular, effectively handling blur, low-light, haze, rain, and snow within a single framework requires models to capture both degradation-shared representations and degradation-specific characteristics.
Therefore, further improvements are still needed to achieve a better balance among restoration fidelity, perceptual quality, computational efficiency, and generalization across complex real-world degradation conditions.

\begin{table*}[t]
\caption{Quantitative results of the LoViF 2026 Second Challenge on Real-World All-in-One Image Restoration.}
\resizebox{\textwidth}{!}{
\begin{tabular}{c|c|c|c|cc}
\hline
~~~~Rank~~~~ & ~~~~Team Name~~~~        & ~~~~Team Leader~~~~  & ~~~~Final Score~~~~ & ~~~~GFLOPs (G)~~~~ & ~~~~Params (M)~~~~ \\ \hline
1    & Re:Pixel & Hongbo Ding       & 42.28       & 131.7          & 1.6          \\
2    & REDnoteMediaLab & Junpeng Jiang       & 41.81       & —          & —          \\
3    & LucidWorld & Xiaodong Zhou       & 41.40       & 8.9          & 1.7          \\
4    & Pheonix & Jingyun Liu       & 34.81       & —          & —          \\
5    & SeeIR & Hongzhen Li       & 34.58       & 869.8          & 174.7          \\
6    & MiVideo & Cheng Li       & 34.52       & 974.22          & 95.95          \\
7    & MiAlgo\_LM & Jiyuan Zhang       & 34.31       & —          & —          \\
8    & Pamy & Xiaofeng Wang       & 33.92       & —          & 28.01          \\
9    & jason0411202 & Yu-Kai Wang       & 32.65       & —          & 271.6          \\
10    & NIT-Oita & Kosuke Shigematsu       & 32.37       & —          & 266.8          \\
11    & ACVLAB & Rong-Lin Jian       & 32.28       & —          & —          \\
12    & MoEFlow & Jialin Zhou       & 32.17       & —          & 28.18          \\
13    & Zzz & Songyu Zhang       & 31.99       & 691.44          & 47.01          \\
14    & BaseLess & E B Benson       & 31.94       & —          & —          \\
15    & O4A & Qirui Chen       & 31.62       & —          & 28.8          \\
16    & GKD\_IR & Xu Zhang       & 31.12       & 1004.48          & 10661          \\ 
17    & IKLab & Saeed Ahmad       & 30.47       & —          & 37.65          \\
18    & sun & Shangquan Sun       & 29.86       & —          & —          \\
19    & Ipara & Behrooz Nobahar-Moghanlou       & 29.25       & —          & —          \\
20    & ustc\_pi\_lab & Ruibo Zhang       & 28.47       & —          & —          \\ 
-    & ULR & Anh-Kiet Duong       & 30.62       & —          & —          \\ \hline
\end{tabular}
}
\label{tab:results_1}
\end{table*}

\section{Teams and Methods}
\label{sec:results}

\subsection{Re:Pixel}

This team proposes ReMamba, a wavelet-driven state-space model for real-world all-in-one image restoration. The network adopts a three-level U-Net-style encoder--decoder and employs the Discrete Wavelet Transform to construct feature representations at $2\times$, $4\times$, and $8\times$ scales. At each level, the input features are decomposed into LL, LH, HL, and HH subbands. ReMamba consists of a High-Frequency State-Aware Block, a Low-Frequency Information Enhancement Block, and a Wavelet Fusion Block. The High-Frequency State-Aware Block combines patch attention with a Vision State Space Module to enhance spatially non-uniform high-frequency degradations, while a gated feed-forward network further refines the resulting features. The patch attention module estimates patch-level restoration difficulty and generates spatial attention weights to guide state-space feature modeling. The Low-Frequency Information Enhancement Block employs a Frequency Matching Transformation to retrieve relevant high-frequency channels and supplement low-frequency representations through dual-branch attention. Finally, the Wavelet Fusion Block combines the high- and low-frequency streams using parallel depthwise convolutions with kernel sizes of $3\times3$ and $5\times5$. Selective Kernel Feature Fusion~\cite{skff2020} is further used to adaptively aggregate the wavelet features. The model is designed to emphasize low-frequency information for haze and low-light restoration while retaining high-frequency structures for blur, rain, and snow removal.

\begin{figure}[t]
\centering
\includegraphics[width=\columnwidth]{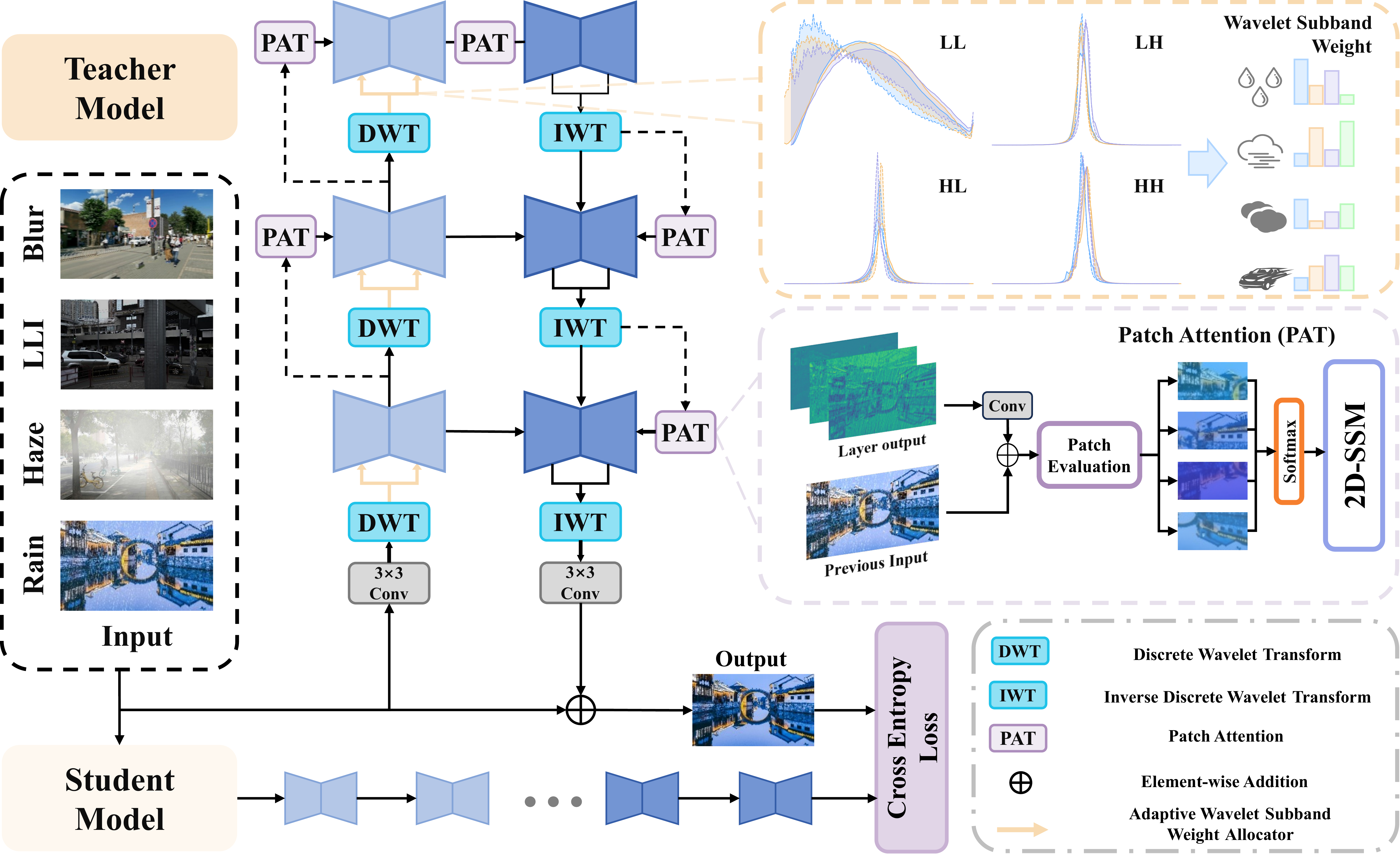}
\caption{The overall architecture of ReMamba proposed by Team Re:Pixel.}
\label{fig:remamba_architecture}
\end{figure}

\noindent \textbf{Training details.}
The model is trained using a two-stage strategy. In the first stage, it is pretrained on the FoundIR~\cite{li2024foundir} and WeatherBench~\cite{guan2025weatherbench} datasets to learn general restoration priors. In the second stage, the pretrained model is fine-tuned on the official LoViF training set, which contains 24,500 paired images with blur, low-light, haze, rain, and snow degradations. The Vision State Space Module is initialized using pretrained MambaIR weights~\cite{mambair2024}. Adam is used as the optimizer with beta values of 0.9 and 0.999. The initial learning rate is 0.0001 and is reduced using cosine annealing over 200 epochs. The batch size is set to 8, and training patches with a spatial size of $256\times256$ are randomly cropped from the original images. Random horizontal flipping, vertical flipping, and rotations by $90^\circ$ are used for data augmentation. The training objective combines L1 reconstruction loss, LPIPS loss with a weight of 0.5, and wavelet-domain frequency consistency loss. A sequential multi-task training strategy with dynamic rehearsal is additionally employed to reduce catastrophic forgetting among different degradation types. The complete model contains approximately 1.60 million parameters and requires 131.7G FLOPs for an input resolution of $1024\times1024$.

\noindent \textbf{Testing details.}
During inference, each $512\times512$ test image is directly processed by ReMamba without image tiling or degradation-specific model selection. The team employs an eight-way self-ensemble strategy consisting of horizontal and vertical flips and rotations by multiples of $90^\circ$. The restored outputs are inversely transformed and averaged to produce the final prediction. No multi-model ensemble is used, and all predictions are generated using a single model checkpoint. The Wavelet Fusion Block performs internal feature fusion between the high- and low-frequency branches. The model consumes less than 2 GB of GPU memory and requires approximately 0.15 seconds to process one image on an NVIDIA RTX 4090 GPU. Their final submission achieves a score of 42.28.

\subsection{REDnoteMediaLab}

This team proposes DreamIR, a pixel-level unified Diffusion Transformer for one-step all-in-one image restoration. Unlike conventional diffusion-based restoration methods that operate in the latent space of a variational autoencoder, DreamIR directly processes RGB pixels to avoid the loss of high-frequency details caused by image encoding and decoding. The model is initialized from HiDream-O1-Image~\cite{cai2026hidream} and unifies four types of inputs into a single token sequence, including text prompt embeddings, semantic features extracted from the degraded image, a diffusion timestep embedding, and image patch tokens. The input image is divided into non-overlapping patches of size 32 by 32, and each flattened patch is projected into a 4096-dimensional feature representation. The backbone contains 32 Transformer decoder blocks and adopts RMSNorm~\cite{zhang2019root}, SwiGLU feed-forward layers~\cite{shazeer2020glu}, grouped-query attention~\cite{ainslie2023gqa}, rotary position embeddings~\cite{su2024roformer}, and query-key normalization. A hybrid attention mechanism preserves the causal prior of the pretrained language model for the text and conditioning tokens while allowing the image tokens to access global information from all conditioning modalities. Instead of iterative denoising, DreamIR learns a direct residual mapping from the degraded image to the clean target and generates the restored image with a single forward pass. The complete model contains approximately 8 billion parameters.

\begin{figure*}[t]
\centering
\includegraphics[width=\linewidth]{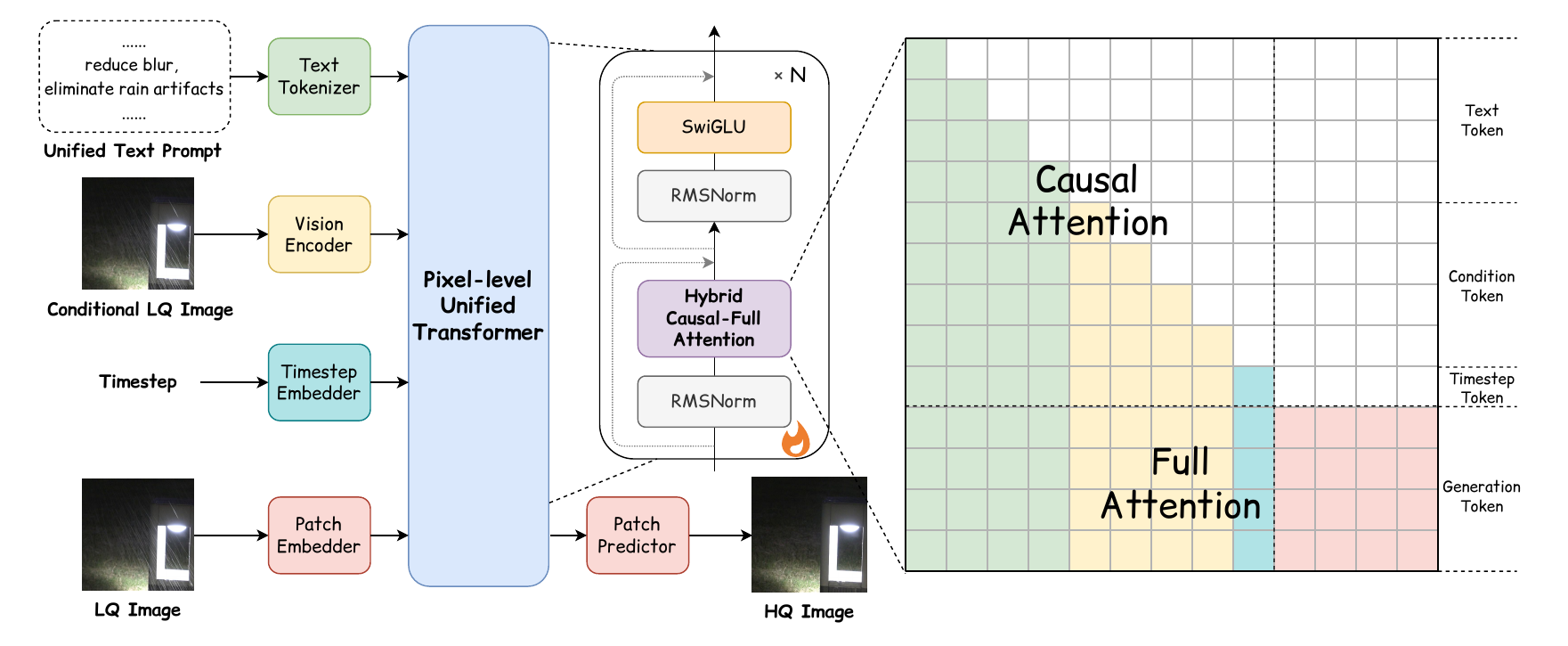}
\caption{The overall architecture of DreamIR proposed by Team REDnoteMediaLab. DreamIR jointly processes text, degraded-image features, timestep information, and image patch tokens to produce a restored image in a single forward pass.}
\label{fig:dreamir_network}
\end{figure*}

\noindent \textbf{Training details.}
DreamIR is trained using a five-stage progressive strategy. In the first stage, the entire model is fine-tuned for 40,000 iterations on FoundIR~\cite{li2024foundir} together with the snow-removal subset of the official competition data. The objective combines flow-matching loss, LPIPS perceptual loss~\cite{zhang2018lpips}, and DINO perceptual supervision~\cite{caron2021emerging}, with weights of 1, 0.1, and 0.01, respectively. In the second stage, the model is fine-tuned for 100 epochs on the official competition training set using only flow-matching supervision. In the third stage, it is further trained for 50 epochs with random rotations and horizontal and vertical flipping applied with a probability of 50 percent. In the fourth stage, frequency-aware refinement is performed for 100,000 iterations using a combination of flow-matching loss and FFT loss, where the FFT term is assigned a weight of 0.1. In the final stage, the model is fine-tuned for 63,000 iterations on an internal dataset using the same loss and augmentation settings as the fourth stage. AdamW is adopted with beta values of 0.9 and 0.999, no weight decay, and a fixed learning rate of 0.00001. ZeRO-3 distributed training is used to partition the model states, and all training stages are conducted using eight NVIDIA A100 GPUs with 40 GB memory each.

\noindent \textbf{Testing details.}
During inference, the team uses the same fixed text prompt for all test images, regardless of degradation type. Each image is restored at a resolution of 512 by 512 through one forward pass without iterative diffusion sampling, test-time augmentation, model ensembling, or category-specific model selection. Inference is performed on a single NVIDIA L20 GPU using bfloat16 precision, and the restored images are saved in JPEG format according to the challenge submission requirements. DreamIR achieves a PSNR of 33.82, an SSIM of 0.8741, an LPIPS value of 0.1513, and an overall testing score of 41.81. The average inference time is approximately 3.86 seconds per image.

\subsection{LucidWorld}

This team proposes a unified framework for real-world all-in-one image restoration based on a hierarchical encoder--decoder architecture. The model adopts a multi-scale U-Net-style backbone and first extracts shallow features using a 3-by-3 convolution. During encoding, the Discrete Wavelet Transform is used for progressive downsampling and decomposes the features into low-frequency structural components and high-frequency detail components. During decoding, the Inverse Wavelet Transform progressively recovers the spatial resolution and reconstructs the restored image. For the low-frequency branch, the team introduces a wavelet-enhanced mixture-of-experts module containing multiple expert branches for modeling global structures, illumination distributions, and large-scale textures under different degradation patterns. A soft routing mechanism adaptively assigns expert weights according to the input degradation. For the high-frequency branch, a dedicated enhancement module restores edges, textures, and fine-grained residual degradations under the guidance of low-frequency structural information. A selective feature fusion mechanism is further employed to aggregate features from different scales and frequency components, allowing the model to jointly recover global structures and local details. The complete network contains approximately 1.7 million parameters and requires about 8.9G FLOPs for an input image of size 256 by 256.

\noindent \textbf{Training details.}
The model is trained using the official LoViF training set together with FoundIR~\cite{li2024foundir}, FoundIR-v2~\cite{chen2026foundir}, and WeatherBench~\cite{guan2025weatherbench}. To alleviate inconsistent convergence among different restoration tasks, the team introduces a dynamic degradation-aware optimization strategy. The degradation types are divided into discrete and continuous groups according to their visual characteristics. Rain and snow are treated as sparse high-frequency particle degradations, while haze and fog are regarded as smooth continuous attenuation degradations. During training, the loss weights of different degradation groups are dynamically adjusted according to the historical trends of their validation losses. A relative preference term increases the contribution of slowly converging or under-optimized degradation groups, while an absolute preference term suppresses unstable groups whose losses oscillate or diverge. The final training objective is obtained by aggregating the restoration losses of different degradation groups using the dynamically estimated weights, resulting in more balanced and stable joint optimization.

\noindent \textbf{Testing details.}
During inference, all degradation types are processed using the same unified restoration model without degradation-specific model selection. The final submission does not use multi-model ensembling or complex test-time fusion. Feature fusion is performed only inside the network through the proposed selective fusion module and the wavelet-based interaction between low- and high-frequency branches. This single-model design reduces inference cost and memory consumption while maintaining unified restoration capability across blur, low-light, haze, rain, and snow degradations. Their final submission achieves a testing score of 41.40.

\subsection{Phoenix}

This team proposes HyRoute, a two-stage mixture-of-experts framework for real-world all-in-one image restoration. The first stage performs degradation-aware restoration through task-specific experts. An image controller adopted from DA-CLIP~\cite{daclip} predicts the degradation category of the input image, and the predicted category is used for hard routing to the corresponding restoration expert. This design allows individual experts to specialize in different degradation patterns and reduces interference among heterogeneous restoration tasks. The second stage introduces a task-shared mixture-of-experts module for content-adaptive refinement. Multiple shared expert branches model complementary restoration behaviors, while a content-aware gating module predicts soft weights according to the visual content of the intermediate features. The outputs of the shared experts are then adaptively combined to exploit common restoration priors across different tasks while preserving flexibility for diverse image structures and local degradation patterns. By combining degradation-aware hard routing with content-adaptive soft expert fusion, HyRoute jointly benefits from task specialization and cross-task knowledge sharing.

\begin{figure}[t]
\centering
\includegraphics[width=\linewidth]{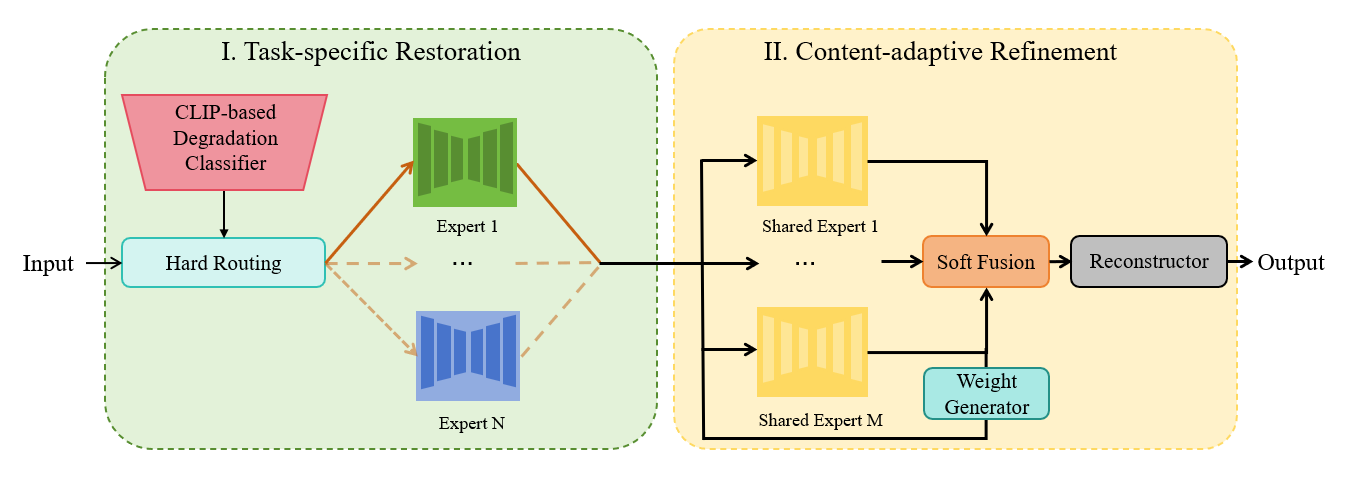}
\caption{The overall framework of HyRoute proposed by Team Phoenix.}
\label{fig:phoenix_hyroute}
\end{figure}

\noindent \textbf{Training details.}
The model is trained using the official LoViF training set together with FoundIR~\cite{chen2026foundir} and WeatherBench~\cite{guan2025weatherbench}. The external datasets are used for pretraining to expose the network to more diverse degradation patterns and restoration scenarios before adaptation to the competition data. The model is optimized with an initial learning rate of 0.0001, which is gradually reduced to 0.000001 using a multistep learning-rate schedule until convergence. Random rotation, horizontal flipping, and vertical flipping are applied as data augmentation to improve robustness and generalization. The factsheet does not provide additional details about the optimizer, batch size, patch size, loss function, or total training iterations.

\noindent \textbf{Testing details.}
During inference, the DA-CLIP image controller first estimates the degradation category of the input image and routes it to the corresponding task-specific expert in the first restoration stage. The resulting features are then processed by the task-shared mixture-of-experts refinement stage, where multiple shared experts are softly weighted and fused according to the image content. The method does not report the use of multi-model ensembling, test-time augmentation, or output-level fusion beyond the internal expert routing and aggregation mechanisms. Their final submission achieves a testing score of 34.81, while the best development score is 34.23.

\subsection{SeeIR}

This team proposes RIRNet, a unified representation-based image restoration network for real-world all-in-one image restoration. RIRNet combines a pretrained ConvNeXt-based DINOv3 backbone~\cite{simeoni2025dinov3} with a multi-scale encoder--decoder restoration branch. The DINOv3 backbone extracts robust high-level semantic representations from the degraded input, while the restoration branch recovers image structures and local details using a feature pyramid encoder and a coarse-to-fine decoder. The restoration blocks adopt the channel-attention mechanism of Restormer~\cite{Zamir2022Restormer} to efficiently model global dependencies in high-resolution features. Multi-scale DINOv3 features are projected to the corresponding encoder dimensions and fused with restoration features through residual addition. The decoder subsequently reconstructs the clean image from coarse to fine scales using bottleneck processing and skip connections. The same network is applied to blur, low-light, haze, rain, and snow images without degradation-specific model switching. The complete model contains approximately 174.7 million parameters, of which 134.6 million belong to the DINOv3 backbone, and requires approximately 869.8 GFLOPs for an input image of size 512 by 512.

\begin{figure*}[t]
    \centering
    \includegraphics[width=\textwidth,height=0.82\textheight,keepaspectratio]{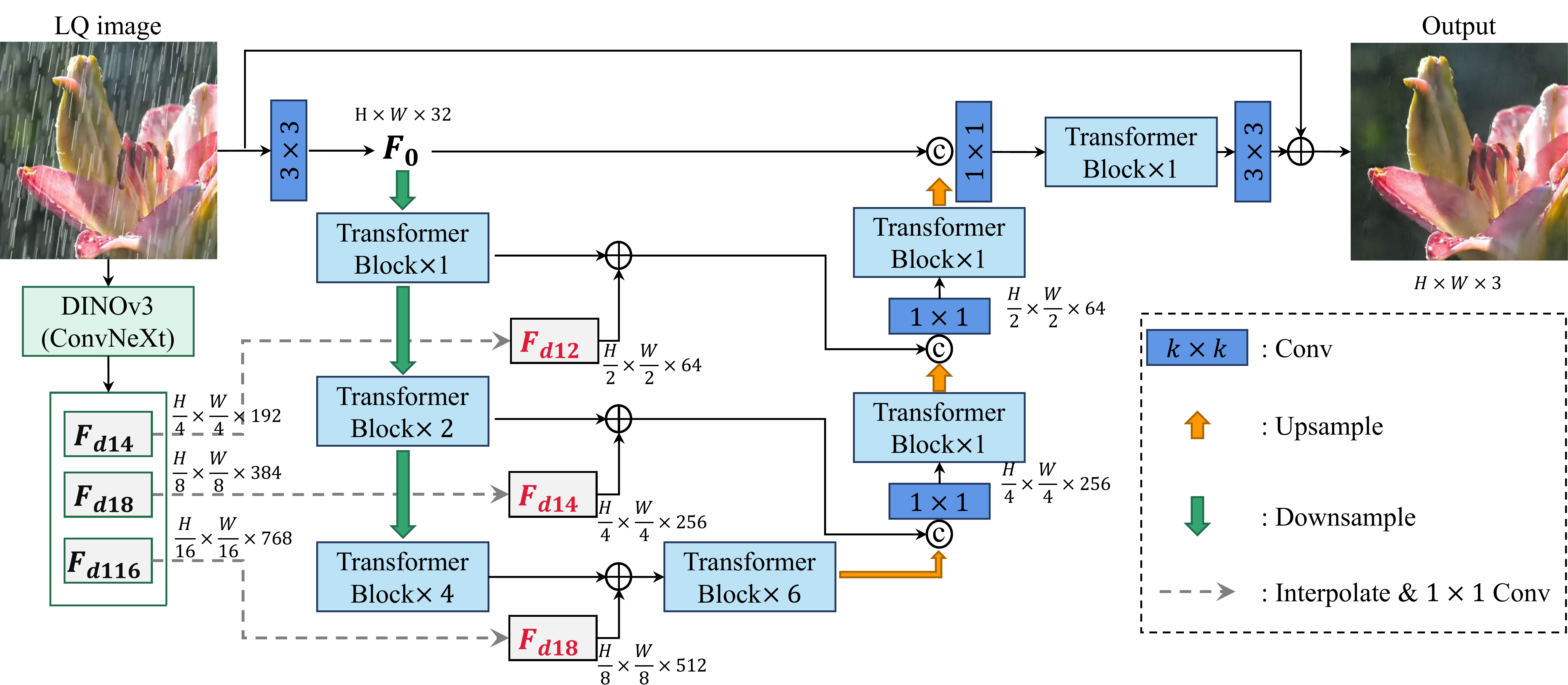}
    \caption{The overall framework of RIRNet proposed by Team SeeIR. A ConvNeXt-based DINOv3 backbone extracts high-level semantic representations, while the multi-scale encoder--decoder branch progressively restores image structures and details.}
    \label{fig:seeir_pipeline}
\end{figure*}

\noindent \textbf{Training details.}
The model is trained using the official LoViF training set together with external data from FoundIR~\cite{li2024foundir} and WeatherBench~\cite{guan2025weatherbench}. The complete external pretraining set contains 310,544 paired images, including 109,480 blur pairs, 93,414 haze pairs, 39,962 low-light pairs, 54,629 rain pairs, and 13,059 snow pairs. To reduce near-duplicate scene bias during fine-tuning, samples sharing the same ground-truth hash are grouped and only one or two pairs are retained from each scene, resulting in 38,656 selected external pairs. Training consists of two stages. The first stage performs external-data pretraining for 400,000 iterations using a progressive patch-size schedule. Patch sizes of 192, 256, and 384 are used for 50 percent, 30 percent, and 20 percent of the iterations, with batch sizes of 40, 25, and 10, respectively. Degradation-balanced sampling is adopted so that each mini-batch contains an equal number of samples from the five degradation categories. The second stage performs mixed-data fine-tuning for 100,000 iterations using patches of size 192 by 192 and a batch size of 36. Each mini-batch contains official LoViF samples and selected external samples at a one-to-one ratio. Adam is used as the optimizer, with initial learning rates of 0.0001 for the restoration network and 0.00001 for the DINOv3 backbone, followed by linear decay to 0.0000001. The model is optimized using only Charbonnier reconstruction loss with an epsilon value of 0.001. No perceptual, adversarial, or frequency-domain losses are employed. FP16 training is conducted on a single NVIDIA RTX 4090 GPU with 48 GB memory.

\noindent \textbf{Testing details.}
During inference, all 500 test images are processed using the same trained RIRNet checkpoint without degradation-specific model selection, checkpoint averaging, output fusion, self-ensemble inference, or category-dependent post-processing. The degraded image is directly passed through the DINOv3 semantic backbone and the multi-scale restoration branch to obtain the final full-resolution output. The reported inference time is approximately 50 milliseconds per image. Their final submission achieves a PSNR of 27.59, an SSIM of 0.85, an LPIPS value of 0.26, and an overall testing score of 34.58.

\subsection{MiVideo}

This team proposes AdaIGFormer, an adaptive illumination-guided Transformer for real-world all-in-one image restoration. The method is developed from RetinexFormer~\cite{cai2023retinexformer} and combines Retinex-inspired illumination decomposition with a U-Net-style Transformer restoration backbone. A task-aware degradation encoder is constructed from a fine-tuned DINOv3 ViT-B/16 model~\cite{simeoni2025dinov3}. The encoder extracts a global condition vector from the degraded image and uses it to dynamically control five parallel processing branches inside each restoration block. These branches include illumination-guided attention, channel attention, overlapping cross-attention adopted from X-Restormer~\cite{chen2023comparative}, channel self-attention adopted from Restormer~\cite{Zamir2022Restormer}, and Fourier-domain processing. A learned gating mechanism predicts the contribution of each branch according to the estimated degradation condition, enabling the network to adapt its feature processing to blur, low-light, haze, rain, and snow. The base feature width is 64, and spatial downsampling is performed using 4-by-4 convolutions with a stride of 2. The complete model contains approximately 95.95 million parameters, of which about 9.85 million are trainable restoration parameters and 86.11 million belong to the fine-tuned and subsequently frozen DINOv3 encoder. The computational cost is approximately 947.22 GFLOPs.

\begin{figure}[t]
\centering
\includegraphics[width=\linewidth]{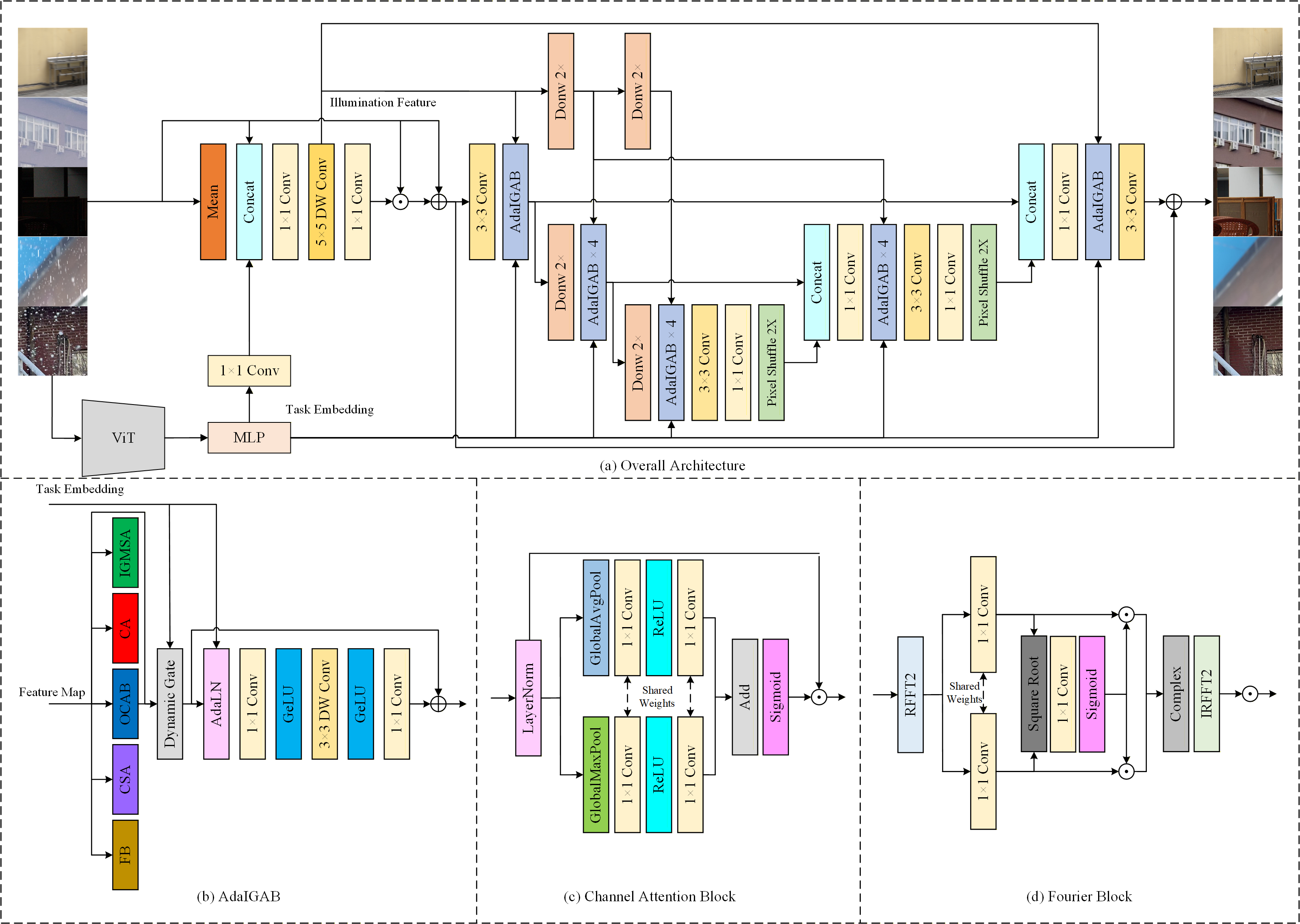}
\caption{The overall architecture of AdaIGFormer proposed by Team MiVideo.}
\label{fig:miVideo_adaigformer}
\end{figure}

\noindent \textbf{Training details.}
The model is trained using the official LoViF training set together with FoundIR, WeatherBench, and RainRAG. Training is conducted on eight NVIDIA A100 GPUs with 80 GB memory, and an exponential moving average of the model parameters is maintained with a decay value of 0.999. The overall training procedure contains six progressive stages. In the first stage, the DINOv3 degradation encoder is fine-tuned on the five LoViF degradation categories using supervised contrastive learning together with an auxiliary classification loss. The encoder produces a 128-dimensional condition embedding and is trained for 2,500 iterations using AdamW with a learning rate of 0.00005. In the second stage, the restoration network is trained on approximately 72,357 image pairs sampled from FoundIR and WeatherBench using patches of size 256 by 256 and a batch size of 16. Muon and AdamW are jointly used with initial learning rates of 0.001 and 0.0003, respectively, and the objective combines Charbonnier loss, LPIPS loss, and SSIM loss with weights of 1.0, 0.1, and 0.3. This stage lasts 150,000 iterations. In the third stage, the patch size is increased to 512 by 512, the batch size is reduced to 4, and training continues for 120,000 iterations. In the fourth stage, 62,194 FoundIR pairs and 24,450 official competition pairs are used, an FFT loss with a weight of 0.01 is introduced, and the model is trained for 200,000 iterations using reduced learning rates. In the fifth stage, 40,770 RainRAG pairs are added, BF16 mixed-precision training is enabled, and training continues for 96,000 iterations with a batch size of 6. In the final stage, the number of bottleneck AdaIGAB blocks is increased from two to four. Previously learned parameters are frozen, and only the newly added blocks are optimized for 20,000 iterations.

\noindent \textbf{Testing details.}
During inference, the degraded image is directly processed by AdaIGFormer, while the frozen DINOv3 encoder first extracts the degradation-aware condition used to control the dynamic multi-branch gates. The restored image is produced by a single model without degradation-specific checkpoint selection. The factsheet does not report the use of test-time augmentation, model ensembling, or output-level fusion. The final checkpoint is selected according to its score on the online leaderboard. The model requires approximately 433 milliseconds to process one image of size 512 by 512 on a single NVIDIA RTX 3080 GPU. Their best testing score is 34.52, while the best development score is 37.13.

\subsection{MiAlgo\_LM}

This team proposes a degradation-aware mixture-of-experts framework that formulates all-in-one image restoration as a perceive-then-restore process. A lightweight degradation router first analyzes the low-quality input and predicts whether it is affected by blur, low-light, haze, rain, or snow. The predicted category and confidence are then used to dispatch the image to a dedicated restoration expert. Instead of using the same backbone for every task, the team selects different restoration models according to the characteristics of each degradation. NAFNet~\cite{chen2022nafnet} is adopted for spatially uniform low-level degradations because of its efficient gated convolutional design, while Restormer~\cite{Zamir2022Restormer} is used for rain removal because its global attention mechanism is better suited to modeling long-range and directional rain streaks. Haze restoration is handled by a two-stage pipeline in which the FLUX.2 4B generative model first reconstructs plausible structures and high-frequency details, followed by a NAFNet refinement model that improves consistency with the input and suppresses generative artifacts. The degradation router is trained to capture degradation-related characteristics rather than semantic scene information, and only geometry-preserving augmentations are retained to avoid changing the original degradation distribution. The final restored output is obtained from the expert branch selected by the router.

\begin{figure}[t]
\centering
\includegraphics[width=0.99\linewidth]{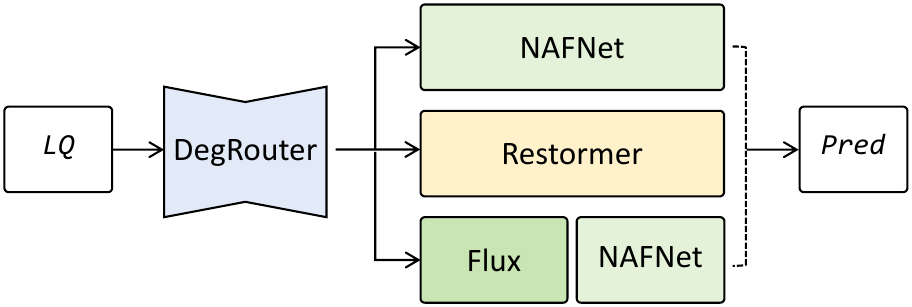}
\caption{The real-world all-in-one restoration framework proposed by Team MiAlgo\_LM.}
\label{fig}
\end{figure}

\noindent \textbf{Training details.}
The expert restoration models are trained using the official LoViF dataset together with FoundIR~\cite{li2024foundir} and WeatherBench~\cite{guan2025weatherbench}. The team employs an LLM-agent-based automatic experiment framework to search for effective training configurations. The framework maintains up to four parallel agents, with each agent operating in an independent code workspace and automatically completing experiment planning, code modification, model training, inference, metric recording, and result submission. Candidate experiments are generated according to previous results, failed configurations, and external references. The automatic search focuses on loss weights, learning-rate schedules, data augmentation, training duration, and checkpoint selection, while the basic NAFNet and Restormer architectures remain unchanged. The factsheet does not report the exact optimizer, batch size, patch size, loss functions, or number of training iterations for the individual expert branches.

\noindent \textbf{Testing details.}
During inference, the degradation router first predicts the degradation category of each test image and selects the corresponding restoration branch. Blur, low-light, and snow images are processed by their assigned discriminative restoration experts, rain images are restored using the Restormer branch, and haze images are processed by the FLUX.2 4B generation stage followed by NAFNet refinement. The factsheet does not report the use of test-time augmentation, checkpoint ensembling, or output averaging. Their final submission achieves an overall testing score of 34.31.

\subsection{Pamy}

This team proposes a Restormer-based framework for real-world all-in-one image restoration. The model retains the four-level encoder--decoder architecture of Restormer~\cite{Zamir2022Restormer} and introduces three modifications to improve restoration across blur, haze, low-light, rain, and snow degradations. First, the original multi-head transposed attention module is replaced with the sparse Top-k attention mechanism from DRSformer~\cite{chen2023learning}. By retaining only the dominant query-key correlations, this design suppresses redundant channel interactions and improves selective feature aggregation under heterogeneous degradations. Second, inspired by X-Restormer++~\cite{Pan2026XRestormerPlusPlus}, the model replaces the conventional global residual connection with a spatially adaptive input scaling mechanism. The network jointly predicts spatially varying input weights and residual correction terms, allowing mildly degraded background regions to be preserved while strongly corrupted regions receive more aggressive restoration. Third, following DSRIR~\cite{WANG2026104380}, multi-scale auxiliary outputs and uncertainty estimation heads are introduced at full, half, quarter, and one-eighth resolutions during training. The estimated uncertainty maps reduce the contribution of unreliable regions and provide additional supervision to intermediate decoder features. These auxiliary branches are discarded during inference, and only the final full-resolution restoration is retained. The complete model contains approximately 28.01 million parameters.

\begin{figure}[t]
\centering
\includegraphics[width=\linewidth]{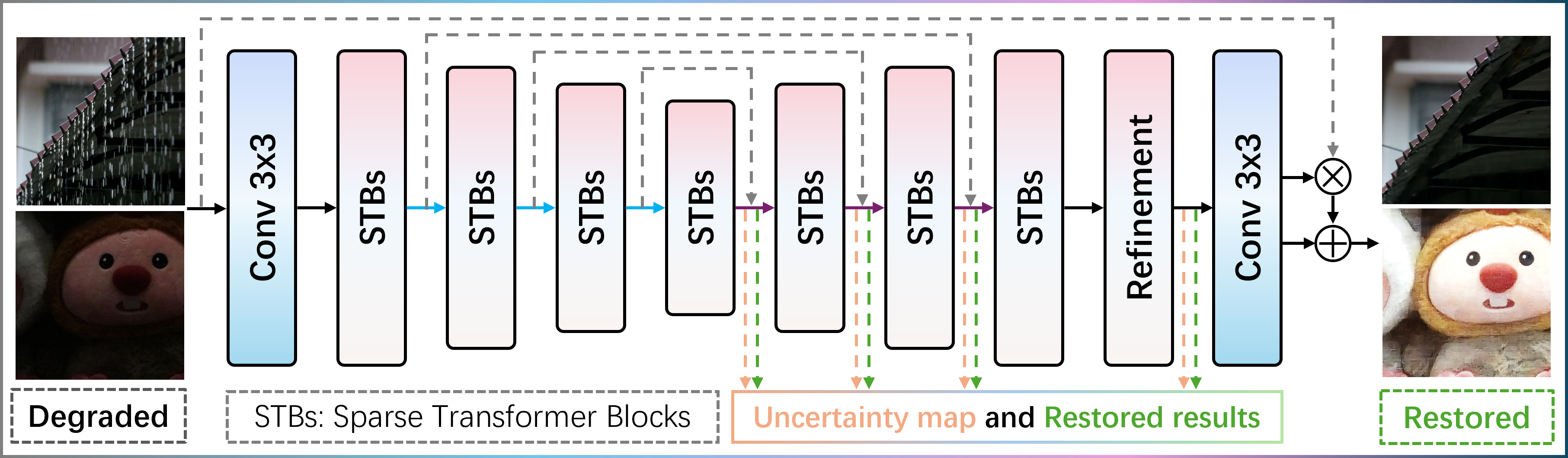}
\caption{The overall framework proposed by Team Pamy.}
\label{fig:pamy_network}
\end{figure}

\noindent \textbf{Training details.}
The model is trained using the official LoViF paired training data together with FoundIR~\cite{li2024foundir} and WeatherBench~\cite{guan2025weatherbench}. Standard data augmentation, including random cropping, horizontal flipping, and rotation, is applied to paired low-quality and ground-truth images. Training is conducted in two stages on a single NVIDIA RTX PRO 6000 GPU. In the first stage, the model is trained for 1,000,000 iterations using patches of size 256 by 256 and a batch size of 5. Adam is adopted with an initial learning rate of 0.0003, beta values of 0.9 and 0.99, and cosine annealing to 0.000001. The objective combines a PSNR-oriented reconstruction loss with multi-scale uncertainty supervision. The auxiliary reconstruction term is assigned a weight of 0.5, while the uncertainty variance regularization term is assigned a weight of 2.0. In the second stage, the model is fine-tuned for 250,000 iterations using patches of size 512 by 512 and a batch size of 1. Adam is again used with a learning rate of 0.00001, beta values of 0.9 and 0.99, and cosine annealing to 0.000001. An FFT loss with a weight of 0.1 is additionally introduced to improve frequency consistency and suppress restoration artifacts.

\noindent \textbf{Testing details.}
During inference, the checkpoint obtained from the second training stage is used for final restoration. Each full-resolution test image of size 512 by 512 is directly processed by the trained network without tiling, test-time augmentation, model ensembling, or degradation-specific model selection. The multi-scale auxiliary outputs and uncertainty heads are used only during training and are removed from the inference pipeline. The final restored image is generated from the full-resolution decoder output. The method requires approximately 0.15 seconds to process one image on an NVIDIA RTX PRO 6000 GPU. Their final submission achieves an overall score of 33.92, with a PSNR of 27.25, an SSIM of 0.81, and an LPIPS value of 0.29.

\subsection{jason0411202}
This team proposes UniNAF-MoRE, a single-checkpoint mixture-of-restoration-experts framework for real-world all-in-one image restoration. Their method employs four restoration experts based on the NAFNet-width64 architecture~\cite{chen2022nafnet}, including three degradation-specific experts for blur, low-light, and haze restoration, and one jointly trained all-in-one expert for rain and snow removal. Each expert contains approximately 67.9M parameters. The weights of all four experts and the routing configuration are packed into a single checkpoint containing 271.6M parameters in total, while only one expert is activated for each input image.
Their submitted method adopts a deterministic degradation-aware routing strategy based on the officially published correspondence between test-image indices and degradation categories. Specifically, blur, low-light, and haze images are processed by their corresponding specialized experts, whereas rain and snow images are restored using the jointly trained all-in-one expert. This routing mechanism introduces no additional trainable parameters and avoids degradation-classification errors during inference. The team also explores a blind router implemented as a linear classifier on frozen DepictQA visual-language features, but this router is not used in the final submission because the index-to-degradation correspondence is publicly available.
The design is motivated by the observation that different degradations benefit from different levels of expert specialization. Their experiments show that a jointly trained model performs effectively on rain and snow because these tasks benefit from cross-degradation knowledge sharing, while blur, low-light, and haze restoration obtain greater improvements from degradation-specific optimization. Compared with their earlier one-step diffusion-based solution, the supervised NAFNet experts achieve considerably higher PSNR and a better overall challenge score.

\begin{figure}[h]
\centering
\resizebox{0.98\linewidth}{!}{%
\begin{tikzpicture}[node distance=4mm and 6mm, font=\small,
  box/.style={draw, rounded corners=2pt, minimum height=7mm, inner sep=4pt, align=center},
  exp/.style={box, fill=blue!8},
  io/.style={box, fill=black!6}]
  \node[io] (in) {Test image\\\texttt{NNNN.jpg} (512$^2$)};
  \node[box, right=of in, fill=orange!12] (route) {Index routing\\(official public mapping,\\no classifier)};
  \node[exp, right=14mm of route, yshift=16mm] (e1) {Blur expert\\NAFNet-w64};
  \node[exp, below=2mm of e1] (e2) {Low-light expert\\NAFNet-w64};
  \node[exp, below=2mm of e2] (e3) {Haze expert\\NAFNet-w64};
  \node[exp, below=2mm of e3] (e4) {AIO expert (Rain/Snow)\\NAFNet-w64};
  \node[box, right=14mm of e2, yshift=-5mm, fill=green!10] (post) {Duplicate-pair patch\\(13 imgs, flag-controlled)\\$+$ DC dither};
  \node[io, right=of post] (out) {JPEG q96\\ss=0, optimize};
  \draw[-{Latex}] (in) -- (route);
  \foreach \e in {e1,e2,e3,e4} \draw[-{Latex}] (route.east) -- (\e.west);
  \foreach \e in {e1,e2,e3,e4} \draw[-{Latex}] (\e.east) -- (post.west);
  \draw[-{Latex}] (post) -- (out);
  \node[draw, dashed, rounded corners, fit=(e1)(e4), inner sep=3pt,
        label={[font=\scriptsize]above:{single checkpoint \texttt{model\_final.pth} (4 experts + routing table)}}] {};
\end{tikzpicture}}
\caption{UniNAF-MoRE inference pipeline. One architecture, one parameter file; the routing signal is the officially published index-to-degradation mapping.}
\label{fig:pipeline}
\end{figure}
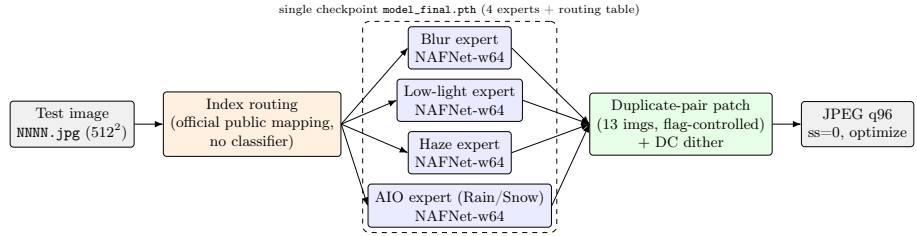

\noindent \textbf{Training details.}
All experts are initialized using the publicly available NAFNet-REDS-width64 pretrained weights. The blur, low-light, and haze experts are independently trained using the corresponding 4,900 paired images from the official LoViF training set. The all-in-one expert is trained using all 24,500 official training pairs and is subsequently used for rain and snow restoration. No additional external training data are employed.
The models are optimized using an $L_1$ reconstruction loss. Training patches of size $256\times256$ are randomly cropped from the original images, with random flipping and $90^{\circ}$ rotation used for data augmentation. AdamW is adopted with $\beta_1=0.9$, $\beta_2=0.9$, and a weight decay of $1\times10^{-4}$. The learning rate is initialized as $2\times10^{-4}$ and gradually reduced to $1\times10^{-7}$ using cosine annealing. The batch size is set to 12, and BF16 automatic mixed precision is employed during training. Each degradation-specific expert is trained for 20,000 iterations, while the all-in-one expert is trained for 40,000 iterations. The best checkpoints are selected according to validation PSNR. For the blur expert, the last 200 blur training pairs are held out for validation. For the remaining experts, 20 images from each relevant degradation category are used as validation samples.

\noindent \textbf{Testing details.}
During inference, each $512\times512$ degraded image is first assigned to an expert according to its test-image index. The selected expert processes the full-resolution image in a single forward pass. The method does not employ image tiling, test-time augmentation, multi-scale inference, or prediction averaging. Although four experts are stored in the same checkpoint, only one expert is activated for each input image.
The restored images are saved in JPEG format with a quality factor of 96, disabled chroma subsampling, and optimized encoding. The inference time is approximately 0.1 seconds per image on one NVIDIA RTX PRO 6000 GPU using BF16 precision and a batch size of one.
The team additionally applies a disclosed duplicate-pair correction to 13 images in the blur subset. These test inputs are identified as nearly pixel-identical to degraded images in the official training set through perceptual hashing, difference hashing, and exact pixel comparison. For these images, the corresponding training ground-truth images are used as outputs. A sparse $+1$ DC perturbation is applied to approximately $1/16$ of the JPEG blocks before encoding. This correction is implemented as an optional post-processing operation and can be disabled in their inference pipeline. Their final submission achieves a challenge score of 32.65, with a PSNR of 25.94, an SSIM of 0.81, and an LPIPS of 0.28.

\subsection{NIT-Oita}

This team proposes a conditional NAFNet for real-world all-in-one image restoration. The model follows the encoder--middle--decoder architecture of NAFNet~\cite{chen2022nafnet} and predicts a residual image that is added to the degraded input. A categorical degradation identifier is provided for each image, covering blur, low-light, haze, rain, and snow. The degradation label is converted into a learned embedding and further processed by a two-layer multilayer perceptron to produce a conditioning vector. This vector modulates the normalized features in every restoration block through learned scale and bias parameters, enabling one network to exhibit degradation-specific restoration behavior. The final model has a base width of 80, encoder block numbers of 2, 2, 4, and 8, 18 middle blocks, decoder block numbers of 2, 2, 2, and 2, and a conditioning dimension of 128. The complete network contains approximately 266.8 million parameters. Compared with training five independent restoration networks, the conditional design shares most of the parameters across degradations while retaining category-specific processing capabilities.

\noindent \textbf{Training details.}
The official LoViF training set contains 24,500 paired low-quality and ground-truth images, with 4,900 pairs for each degradation category. The final model is obtained through a four-stage training strategy. In the first stage, the width-80 conditional NAFNet is trained from scratch using the official optimization split together with 2,103 GoPro blur and sharp image pairs~\cite{nah2017deep}. In the second stage, the model is trained only on the official LoViF data to reduce the domain gap introduced by the external blur dataset. In the third stage, continuation training is performed using the official data together with GoPro, 6,397 HIDE blur and sharp pairs~\cite{shen2019human}, and 1,021 SICE low-light pairs~\cite{cai2018learning}. In the final stage, the model is again trained using only the official LoViF optimization split to recover competition-domain performance. The final stage contains 60,000 optimization steps with a batch size of 12 and a patch size of 256. AdamW is used with beta values of 0.9 and 0.99 and a weight decay of 0.0001. The learning rate follows a cosine schedule from 0.00002 to 0.0000001, with 1,000 warm-up steps. The training objective combines Charbonnier reconstruction loss and frequency-domain FFT loss, with the FFT loss assigned a weight of 0.05. Gradient clipping is set to 1.0, and exponential moving average weights are tracked using a decay value of 0.999. However, the raw model weights rather than the exponential moving average weights are used in the final submission.

\noindent \textbf{Testing details.}
During inference, the degradation category is determined from the competition-defined test-image order. Images numbered from 0001 to 0100 are treated as blur, images from 0101 to 0200 as low-light, images from 0201 to 0300 as haze, images from 0301 to 0400 as rain, and images from 0401 to 0500 as snow. Each image is processed using a single forward pass of the conditional NAFNet with the corresponding degradation label. The method does not use test-time augmentation, model ensembling, checkpoint averaging, test-time adaptation, or output fusion. Restored images are saved in JPEG format with a quality value of 96 and 4:4:4 chroma sampling. The final submission uses the raw \texttt{exp014} checkpoint and achieves an overall score of 32.37, with a PSNR of 25.75, an SSIM of 0.81, an LPIPS value of 0.29, and an average runtime of approximately 0.1 seconds per image on a GPU.

\subsection{ACVLAB}

This team proposes DORNet, a degradation-oriented residual network with semantic verification for real-world all-in-one image restoration. DORNet processes blur, haze, rain, snow, and low-light images using a single restoration checkpoint. The model follows an encoder--decoder architecture and separates degradation modeling, restoration feature extraction, residual correction, semantic preservation, and skip-feature calibration into different components. DSE-Dyn first estimates the degradation category, severity, and restoration policy from the degraded input. These predictions do not select different restoration models but instead control the internal behavior of the same network. The main restoration backbone adopts MIRAGE-style blocks~\cite{mirage}, which combine global context modeling, local convolutional processing, and channel-statistic calibration. An Adaptive Degradation Expert Correction module then performs spatially adaptive residual correction through one shared expert and several specialized experts. A feature router predicts pixel-level expert weights according to the degradation prior, local high-frequency statistics, and restoration features. To prevent the residual correction from damaging image content, the team introduces a semantic feature verifier based on a frozen DINOv3 ConvNeXt backbone~\cite{simeoni2025dinov3,convnext}. Multi-scale semantic features are used to evaluate whether the predicted residual is consistent with the structures and objects in the input image, and unsafe corrections are suppressed through a learned spatial gate. The network further employs WaveSkip, which decomposes encoder skip features into low- and high-frequency wavelet components and calibrates the reliability of the high-frequency components before decoder fusion. In this way, rain streaks, snow particles, blurred edges, and other unreliable high-frequency signals can be suppressed while useful structural details are retained.

\noindent \textbf{Training details.}
The model is trained using the official LoViF training set together with FoundIR~\cite{li2024foundir} and WeatherBench~\cite{guan2025weatherbench}. A staged training strategy is adopted. The network is first trained on the mixed official and external paired restoration datasets to learn general multi-degradation restoration priors and is subsequently fine-tuned on the official LoViF data to reduce the domain gap. AdamW is used as the optimizer with cosine learning-rate decay. The total training length is 300,000 iterations, and the learning rate ranges from 0.0002 to 0.00002 according to the training stage. The batch size is set to 8, and image patches of size 256 by 256 are randomly cropped during training. Data augmentation includes random horizontal flipping, vertical flipping, and rotation. The main objective combines L1 reconstruction loss and frequency-domain FFT loss. The frozen DINOv3 ConvNeXt backbone is used only to provide semantic and structural priors and does not receive restoration gradients.

\noindent \textbf{Testing details.}
During inference, all test images are processed using the same DORNet checkpoint without degradation-specific model selection, filename-based routing, checkpoint ensembling, or multiple restoration networks. DSE-Dyn estimates the degradation-aware restoration policy, while the MIRAGE backbone, adaptive residual experts, semantic verifier, and WaveSkip modules operate jointly inside the single network. The expert routing and semantic fusion mechanisms are feature-level operations rather than output-level ensembles. Their final submission achieves an overall testing score of 32.28, with a PSNR of 25.67, an SSIM of 0.81, and an LPIPS value of 0.30.

\subsection{MoEFlow}

This team proposes a statistics-conditioned mixture-of-experts rectified-flow framework for real-world all-in-one image restoration. The core network is a four-level conditional U-Net with channel widths of 48, 96, 192, and 384. The shallow stages employ conditional NAF-style blocks, while the deeper stages use Restormer-style attention blocks to enlarge the effective receptive field. The restoration process is conditioned on three types of information, including a rectified-flow time embedding, a degradation-category embedding, and a 28-dimensional image-statistics vector extracted from the degraded input. These conditioning signals are fused through multilayer perceptrons and injected into the restoration backbone. Mixture-of-experts adapters are inserted into the deeper encoder, bottleneck, and decoder stages, using eight experts and top-two routing. Gated skip connections and a multi-scale task aggregator are further introduced to adaptively combine low-level spatial details and high-level task information. The model learns a one-step rectified-flow prediction that estimates the restoration residual from the degraded input and directly produces the restored image in a single forward pass. The complete network contains approximately 28.18 million parameters.

\noindent \textbf{Training details.}
The main model is trained using only the official paired LoViF training data, without external paired restoration datasets. AdamW is adopted with an initial learning rate of 0.00005, a weight decay of 0.0001, a crop size of 192, four sampled flow times for each training pair, and gradient clipping with a maximum norm of 1.0. Paired horizontal and vertical flipping are applied with a probability of 0.5, and automatic mixed precision is disabled. The training objective combines smooth L1 flow loss, Charbonnier endpoint reconstruction loss, Laplacian edge loss, frequency-domain loss, luminance loss, structural similarity loss, LPIPS perceptual loss, and an expert-balance loss. Their corresponding weights are 20.0, 0.5, 0.08, 0.10, 0.20, 0.05, 0.10, and 0.001, respectively. The LPIPS term uses a frozen AlexNet backbone. The team further performs transductive pseudo-label fine-tuning using the official test inputs and restored results generated by their own base model. This pseudo-label stage uses a learning rate of 0.00001, repeats pseudo samples 15 times, assigns the pseudo loss a weight of 0.25, and performs 3,000 training batches per epoch. Since pseudo labels may introduce test-set bias, the pseudo-finetuned model is used only as a lightweight component in the final output fusion.

\noindent \textbf{Testing details.}
During inference, the degradation category is determined according to the competition-defined image order, and each input is restored using one-step rectified-flow prediction. The final system employs flip-based test-time augmentation, checkpoint ensembling, pseudo-finetuned model fusion, and category-aware lightweight blending. The base prediction combines three checkpoints using weights of 60 percent, 20 percent, and 20 percent. For rain and snow images, the final result combines 85 percent of the base prediction with 15 percent of the pseudo-finetuned prediction. For blur, low-light, and haze images, the final result combines 75 percent of the base prediction, 20 percent of the pseudo-finetuned prediction, and 5 percent of an additional short-training checkpoint. The restored images are saved as RGB JPEG files with a quality value of 96. Their final submission achieves an online score of 32.17.

\subsection{Zzz}

This team proposes a classification-guided restoration system consisting of a DATPRL-IR-based all-in-one restoration model, a blur-specific NAFNet expert, and a MobileNetV3-Large degradation classifier. The all-in-one branch follows a Restormer-style encoder--decoder architecture and retains only the task prompt pool from DATPRL-IR~\cite{datprl_ir}, removing the original domain prompt pool because the challenge focuses on natural image restoration. The task prompts provide degradation-aware representations for the restoration backbone, while a prompt-guided feature routing module is introduced into the decoder to adaptively fuse multi-level encoder features. The encoder features are first aligned to the target decoder resolution and channel dimension, and the task prompts are then used to generate soft routing weights for feature aggregation. During inference, MobileNetV3-Large~\cite{mobilenetv3} predicts the probabilities of blur, haze, low-light, rain, and snow. Inputs classified as blur with sufficiently high confidence are restored by the NAFNet expert~\cite{chen2022nafnet}, whereas the remaining inputs are processed by the DATPRL-IR-based all-in-one model. For high-confidence haze inputs, the team additionally applies mild contrast-limited adaptive histogram equalization to the luminance channel in the LAB color space and blends the enhanced result with the original input before restoration. The complete system contains approximately 47.01 million parameters and requires about 691.44 GFLOPs.

\begin{figure*}[t]
\centering
\includegraphics[width=0.95\textwidth]{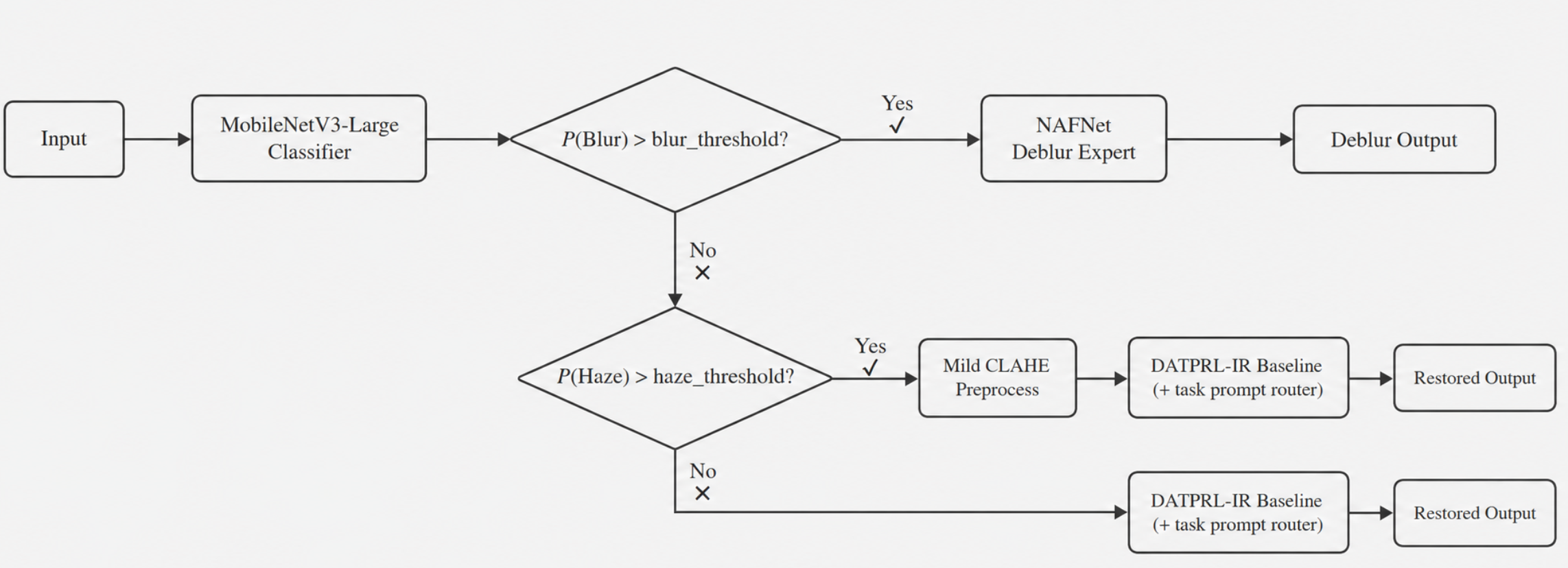}
\caption{The overall framework of Team Zzz. The degradation classifier routes high-confidence blur inputs to the NAFNet expert, while the remaining inputs are restored by the DATPRL-IR-based all-in-one model. Mild CLAHE preprocessing is applied to high-confidence haze inputs.}
\label{fig:zzz_pipeline}
\end{figure*}

\noindent \textbf{Training details.}
The system is trained in three stages using only the official LoViF training data. In the first stage, the modified DATPRL-IR model is trained for 50 epochs using patches of size 128 by 128 and a batch size of 10, followed by another 20 epochs using patches of size 256 by 256 and a batch size of 6. During the second phase, the sampling frequency of blur images is reduced because the blur-specific expert handles most deblurring cases, while the sampling frequencies of the remaining degradations are increased. In the second stage, the NAFNet expert is initialized using the publicly available NAFNet-SIDD-width32 pretrained weights and fine-tuned for 100 epochs on the official blur subset using full-resolution patches of size 512 by 512 and a batch size of 8. In the third stage, MobileNetV3-Large is initialized using ImageNet pretrained weights and trained as a five-class degradation classifier for 50 epochs using patches of size 512 by 512 and a batch size of 32. AdamW is used for all three components. The restoration models use a cosine learning-rate schedule from 0.0001 to 0.000001, while the classifier uses a cosine schedule from 0.001 to 0.00001. The DATPRL-IR model and NAFNet expert are trained using a combination of Charbonnier loss and LPIPS loss~\cite{zhang2018lpips}, whereas the degradation classifier is optimized using cross-entropy loss.

\noindent \textbf{Testing details.}
During inference, the degradation classifier first determines whether an image should be processed by the blur-specific NAFNet expert or the DATPRL-IR-based all-in-one branch. High-confidence haze images in the all-in-one branch receive mild CLAHE preprocessing before restoration. The team further employs four-way test-time augmentation using the original image, horizontal flipping, vertical flipping, and combined horizontal and vertical flipping. Each transformed image is restored independently, inverse-transformed to the original orientation, and combined by taking the pixel-wise median of the four predictions. The complete system requires approximately 0.351 seconds to process one image on an NVIDIA V100 GPU with 32 GB memory. Their final submission achieves a development score of 32.56 and a testing score of 31.99.

\subsection{BaseLess}

This team proposes FFC-Restore, a unified mixture-of-experts network for real-world all-in-one image restoration. The method adopts a U-Net-style encoder-decoder architecture with three feature scales and channel widths of 64, 128, and 256. Given a degraded RGB image, the network predicts a residual image and adds it to the input to obtain the restored result. The encoder contains conventional residual blocks and FourierResidualBlocks constructed using Fourier Feature Convolution layers~\cite{FFC}. Each Fourier Feature Convolution combines a local convolution branch with a frequency-domain branch based on the real fast Fourier transform, learned amplitude modulation, and inverse Fourier transform. PixelAttention modules are used to adaptively modulate the extracted features. At the lowest-resolution bottleneck, the network introduces a mixture-of-experts module containing two experts. The first expert combines conventional residual processing and Fourier-domain processing to restore geometric structures and weather-related artifacts. The second expert mainly uses FourierResidualBlocks to handle frequency and illumination variations. A learned router predicts soft weights to combine the outputs of the two experts. The decoder uses PixelShuffle upsampling and gated skip connections before the final reconstruction layer. The same model is used for blur, low-light, haze, rain, and snow restoration without requiring degradation labels during inference.

\begin{figure}[t]
\centering
\includegraphics[width=\linewidth]{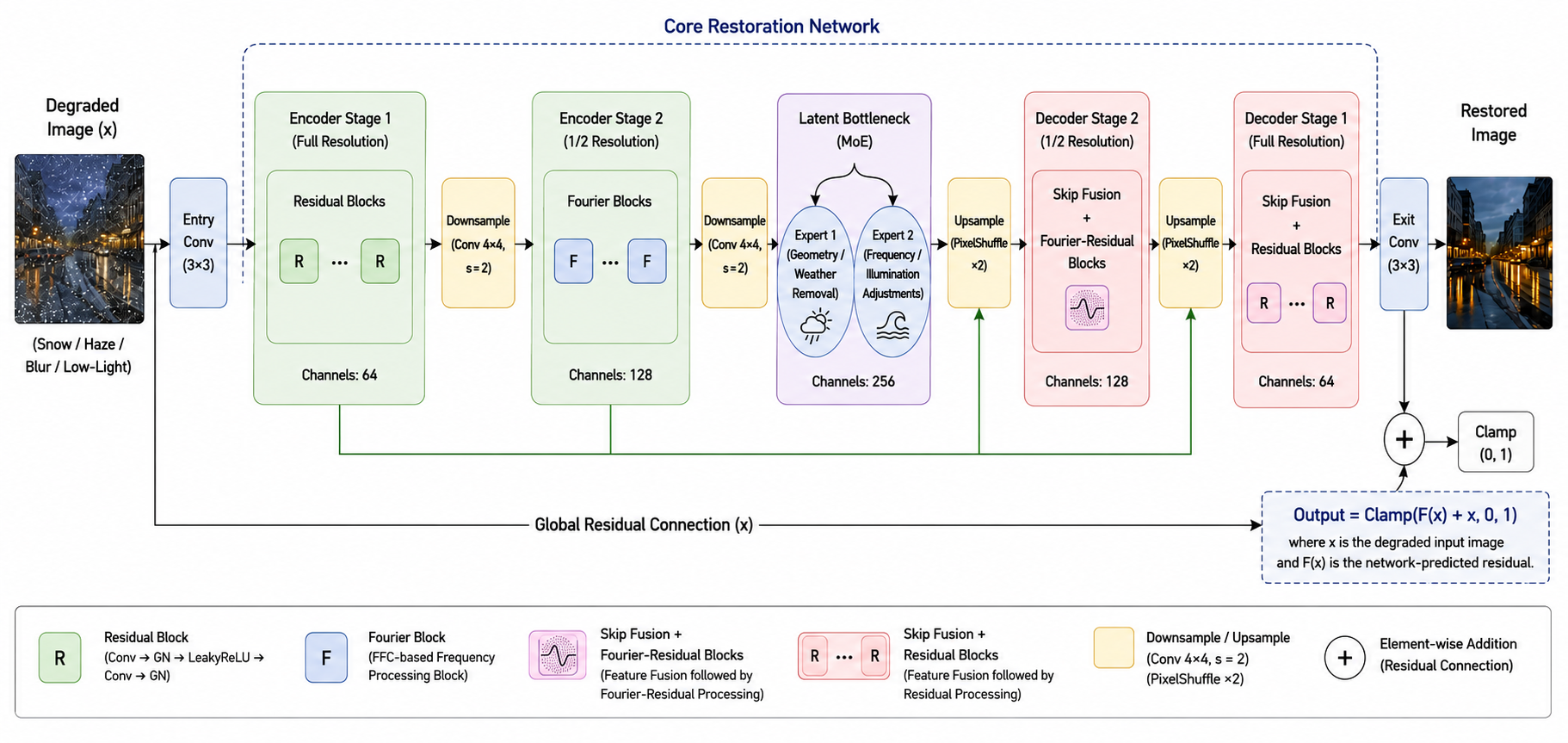}
\caption{The overall framework of Team BaseLess. FFC-Restore combines an encoder-decoder architecture, Fourier Feature Convolution blocks, pixel attention, and a mixture-of-experts bottleneck for unified image restoration.}
\label{fig:baselesspipeline}
\end{figure}

\noindent \textbf{Training details.}
The model is trained using the official LoViF paired training data together with a subset of the GGT-100K dataset~\cite{GGT100k}. All degradation categories are merged into a unified training set containing paired low-quality and ground-truth images. The main restoration network is trained without loading pretrained restoration weights. AdamW is used as the optimizer with an initial learning rate of 0.0002 and a weight decay of 0.0001. Cosine annealing is applied over 80 epochs, and the final learning rate is set to 0.000001. The batch size is 32, and image patches with a spatial size of 256 by 256 are randomly cropped for training. Horizontal and vertical flipping are applied as data augmentation, and gradient clipping is used to stabilize optimization. The training objective combines Charbonnier loss, structural similarity loss, LPIPS loss, and a boundary-weighted reconstruction loss. The LPIPS network uses a frozen AlexNet backbone. The final submitted checkpoint is obtained after 80 epochs.

\noindent \textbf{Testing details.}
During inference, all degraded images are processed by the same UnifiedMoERestorer without degradation labels, degradation classifiers, or category-specific checkpoints. Each input image is padded by reflection so that its height and width are divisible by eight. After restoration, the output is cropped back to the original image size and saved in JPEG format with a quality value of 96. The method does not use model ensembling, test-time augmentation, multi-model fusion, or degradation-specific output selection. The two internal experts are combined through the learned soft routing mechanism. Their final submission achieves a score of 31.94 in the testing phase, while the best development-phase score is 31.55.

\subsection{O4A}

This team proposes a blind all-in-one restoration network based on AdaIR~\cite{cui2025adair}. The model adopts a four-scale U-Net encoder--decoder containing approximately 28.8 million parameters. Each scale contains Transformer blocks that combine window-based attention with depthwise-convolutional feed-forward networks, while skip connections preserve high-resolution details for decoder reconstruction. Frequency-Adaptive Modules are inserted at all feature scales to dynamically modulate low- and high-frequency information. Each module transforms the features into the frequency domain and learns a soft mask that controls the contribution of different frequency bands. This enables the same network to handle rain and snow patterns dominated by high-frequency components, haze and low-light degradations associated with low-frequency information, and blur affecting a wider frequency range. The method does not require degradation labels, classifiers, task-specific checkpoints, or expert switching, and all five degradation categories are restored using the same model parameters.

\noindent \textbf{Training details.}
The model is trained using only the 24,500 official LoViF paired training images, without external datasets or pretrained restoration checkpoints. The training set contains 4,900 pairs for each of the five degradation categories, and a balanced sampler ensures that every mini-batch contains an equal number of samples from each category. The model is trained for 20 epochs using patches of size 384 by 384 on seven NVIDIA A100 GPUs with 80 GB memory. AdamW is adopted with an initial learning rate of 0.00002, a cosine learning-rate schedule, and a two-epoch linear warm-up. Gradient clipping is set to 0.5, and gradient accumulation over two iterations produces an effective batch size of 70. BF16 mixed-precision and distributed data-parallel training are employed. The training objective combines Charbonnier reconstruction loss, luminance-channel structural similarity loss, and frequency-domain reconstruction loss, with weights of 1.0, 0.5, and 0.1, respectively. An exponential moving average of the model parameters is maintained with a decay value of 0.999, and the exponential-moving-average checkpoint is used for final inference.

\noindent \textbf{Testing details.}
During inference, each input image is reflection-padded so that its height and width are divisible by 16. The team employs an eight-way self-ensemble based on rotations and horizontal flipping. Each transformed image is restored independently and transformed back to its original orientation, after which the eight predictions are combined using a pixel-wise median. The median aggregation is used instead of averaging to reduce the influence of isolated restoration artifacts. The restored images are saved as JPEG files with a quality value of 96, disabled chroma subsampling, and optimized encoding. The method requires approximately 3 seconds per image on an NVIDIA A100 GPU with self-ensemble and approximately 0.4 seconds without self-ensemble. Their final submission achieves an overall testing score of 31.62, with a PSNR of 25.0, an SSIM of 0.81, and an LPIPS value of 0.29.

\subsection{GKD\_IR}

This team proposes BioIR, a bio-inspired framework for real-world all-in-one image restoration. BioIR adopts a U-shaped encoder--decoder architecture and uses the BioModule as its main feature-mixing component~\cite{cui2026bio}. Inspired by the interaction between foveal and peripheral vision in the human visual system, the BioModule contains two complementary pathways. The Peripheral-to-Foveal pathway pools local features into region-level descriptors and propagates large-receptive-field contextual information back to individual spatial locations through pixel-to-region affinity. The Foveal-to-Peripheral pathway preserves fine-grained structures by combining element-wise feature modulation with dynamic convolution. The outputs of the two pathways are recalibrated through element-wise multiplication, enabling high-order interaction between global context and local details within a lightweight convolutional design. To adapt the original BioIR architecture to the all-in-one restoration setting, the team further introduces a degradation-aware prompt interaction mechanism. A prompt generator learns multiple degradation components and predicts input-dependent prompt weights from global feature representations. The resulting prompts are injected into the bottleneck and the second decoder stage, allowing the network to adapt its restoration behavior to blur, haze, low-light, rain, and snow degradations.


\noindent \textbf{Training details.}
The model is trained using the official competition data without additional external training datasets. Adam is adopted as the optimizer, and the training objective combines L1 reconstruction losses calculated in both the spatial and frequency domains. The model is trained for 200 epochs using image patches of size 256 by 256, a batch size of 1, and an initial learning rate of 0.0002. Random horizontal and vertical flipping are applied as data augmentation. The degradation-aware prompt generator and the main restoration network are optimized jointly so that the learned prompts can represent different degradation characteristics and guide the restoration process at multiple network stages.

\noindent \textbf{Testing details.}
During inference, each degraded image is directly processed by a single BioIR checkpoint without degradation-specific model selection or explicit multi-model ensembling. The foveal and peripheral feature streams are fused internally through the BioModule, while the degradation-aware prompts dynamically adapt the restoration features to the input degradation. The inference implementation supports checkpoint loading, restored-image saving, NIQE evaluation, and basic efficiency measurement. The model contains approximately 15.86 million parameters and requires about 1004.48 GFLOPs. It produces approximately 10,661 million activations, consumes about 1,589 MB of GPU memory, and requires approximately 439 milliseconds per image on an NVIDIA RTX 3090 GPU. Their final submission achieves an overall score of 31.12.

\subsection{IKLab}

This team proposes a router-gated mixture-of-experts framework for blind multi-degradation image restoration. The main restoration backbone is adapted from PromptCIR~\cite{li2024promptcir} and PromptIR~\cite{potlapalli2023promptir}, retaining their U-Net-style encoder--decoder and prompt-interaction design while replacing the original window-attention blocks with Mamba-based state-space blocks~\cite{gu2023mamba,liu2025dpmambair}. A lightweight degradation router predicts a soft distribution over blur, low-light, haze, rain, and snow directly from the input image. The predicted distribution conditions the prompt modules at multiple decoder scales and also controls a frequency-domain deblurring branch. A Haar-wavelet refinement block is inserted before the output head to enhance frequency consistency. The single backbone contains approximately 37.65 million parameters and can independently restore all five degradation types. For the final submission, the team further introduces two router-gated specialist branches. An EVSSM model~\cite{kong2025evssm} is used for blur restoration, while a low-light branch combines DarkIR~\cite{feijoo2025darkir} with two independently fine-tuned HVI-CIDNet models~\cite{yan2025hvi}. A content-aware hard gate combines the router probabilities with image brightness and Laplacian variance to select the backbone output, the blur expert, or the low-light expert for each image.

\begin{figure}[t]
    \centering
    \includegraphics[width=\linewidth]{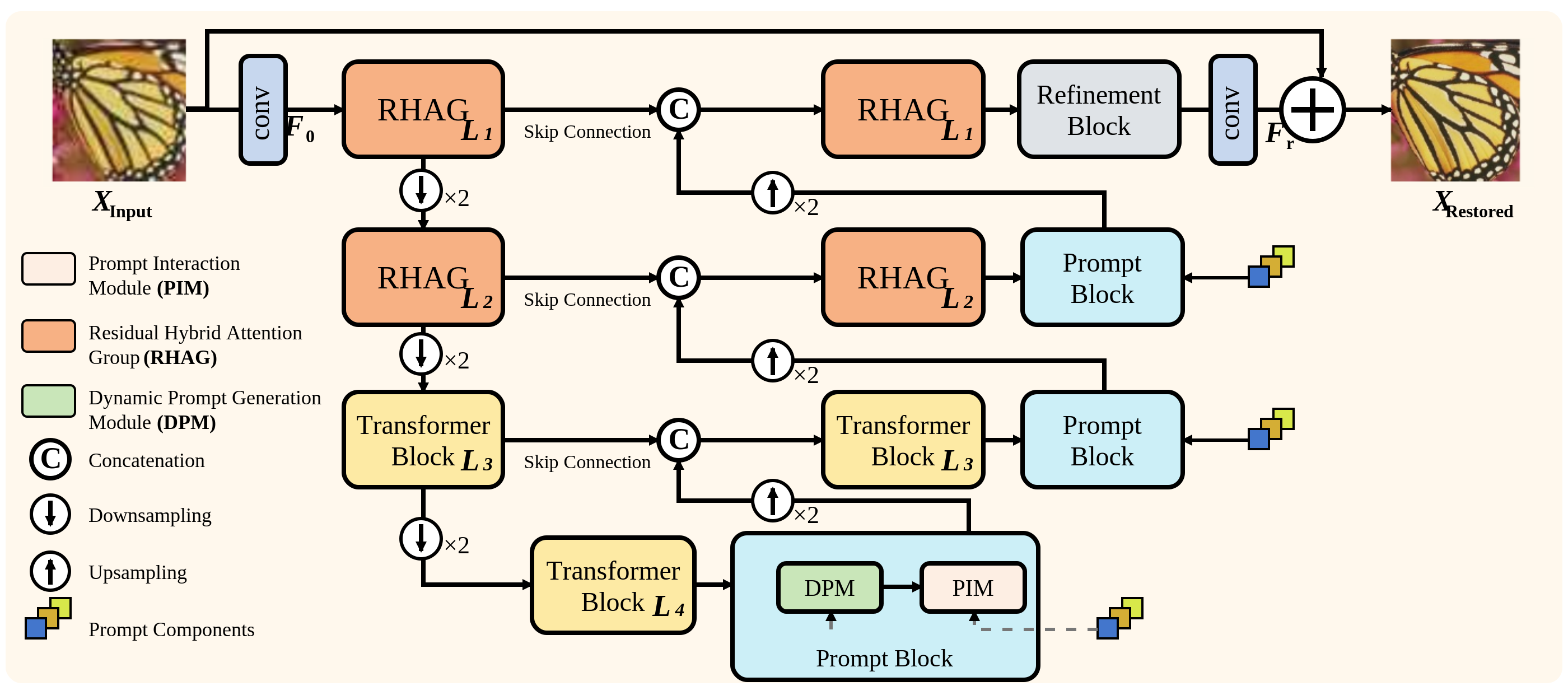}
    \caption{The PromptCIR-style backbone adopted by Team IK Lab. The original attention blocks are replaced with Mamba-based state-space blocks, while degradation routing, routed prompts, wavelet refinement, and a gated deblurring branch are added to the restoration network.}
    \label{fig:iklab-backbone}
\end{figure}

\begin{figure*}[t]
    \centering
    \includegraphics[width=0.92\linewidth]{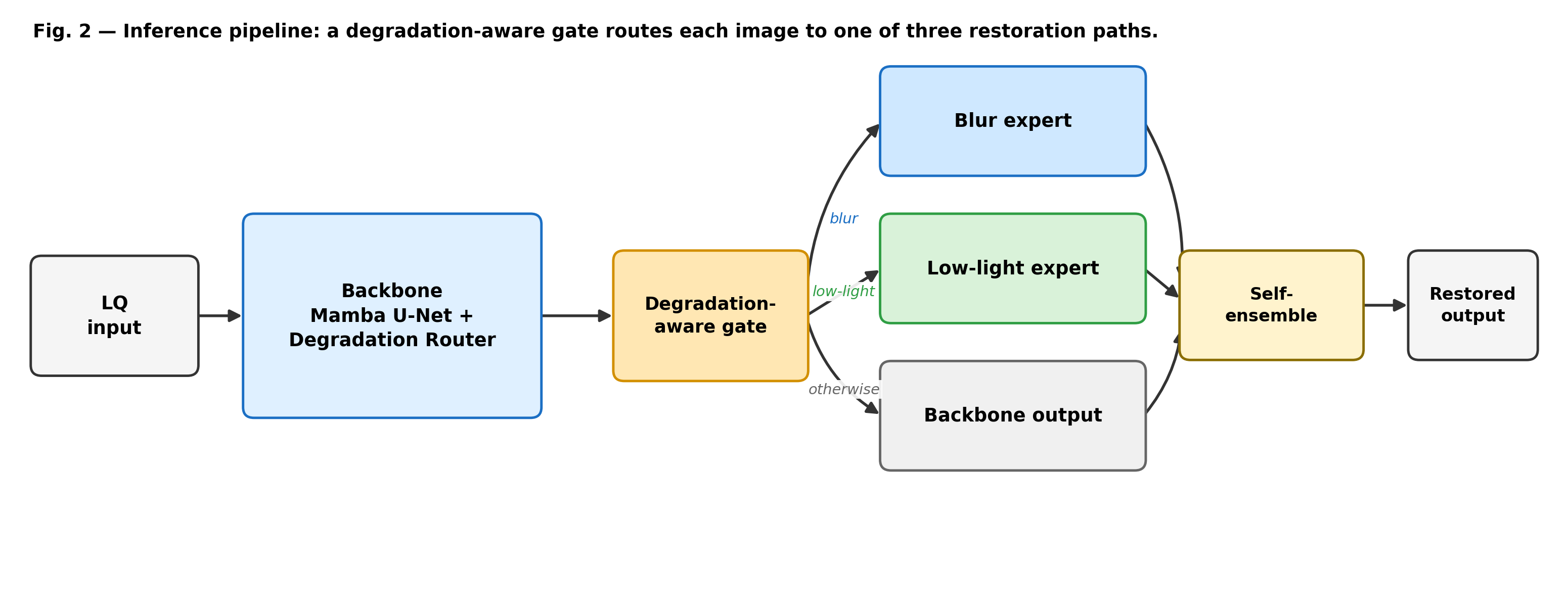}
    \caption{The final inference pipeline of Team IK Lab. The backbone predicts both the restored image and degradation probabilities, after which a content-aware gate selects the backbone result, the EVSSM blur expert, or the fused low-light expert.}
    \label{fig:iklab-pipeline}
\end{figure*}

\noindent \textbf{Training details.}
The main backbone and all specialist models are fine-tuned using the official LoViF paired training data. No additional paired restoration dataset is directly added to the backbone training set, although publicly released pretrained checkpoints are used to initialize the specialist models. The backbone is optimized using AdamW with beta values of 0.9 and 0.99, a weight decay of 0.0001, linear warm-up, cosine annealing, exponential moving average weights, and FP16 mixed-precision training. The objective combines L1 loss, Charbonnier loss, FFT loss, multi-scale structural similarity loss, and an auxiliary cross-entropy loss for degradation classification. Higher task weights are assigned to blur, low-light, and haze because these categories are more difficult on the internal validation split. Progressive patch-size training and multiple cosine warm-restart stages are employed for approximately 50,000 iterations. The EVSSM blur expert is initialized from a publicly released RealBlur-pretrained checkpoint and fine-tuned for approximately 60,000 iterations using official blur pairs together with synthetic blur generated from the clean targets of all five tasks. Its objective combines spatial-domain and frequency-domain L1 losses. DarkIR is initialized from a public LOL-v2-pretrained checkpoint, while the two HVI-CIDNet models are initialized from public cross-dataset checkpoints and fine-tuned on the official low-light subset. DarkIR and HVI-CIDNet are trained with reconstruction, frequency, and structural similarity losses, and HVI-CIDNet is trained using full precision to avoid numerical instability in its color-space transformations.

\noindent \textbf{Testing details.}
During inference, the backbone first produces a restoration result and a five-category degradation distribution. A content-aware gate then combines the predicted blur and low-light probabilities with mean brightness and Laplacian variance to select one of three outputs. Blur images with sufficient confidence are processed by the EVSSM expert, low-light images are processed by a fusion of DarkIR and two HVI-CIDNet models, and the remaining images retain the backbone output. The low-light result is obtained using fixed weights of 50 percent for DarkIR and 25 percent for each HVI-CIDNet model, followed by mild unsharp-mask enhancement. Every branch independently applies eight-view geometric self-ensemble using rotations and horizontal flips, and the transformed predictions are combined using the pixel-wise median. On the official 500-image test set, 69 images are routed to EVSSM, 98 images are routed to the low-light fusion, and 333 images use the backbone output. The average runtime is approximately 3.21 seconds per image on a single NVIDIA RTX 4090 GPU, with a peak memory consumption of approximately 8.8 GB. Their final submission achieves an overall testing score of 30.4731, with a PSNR of 24.1218, an SSIM of 0.7983, and an LPIPS value of 0.3264.

\subsection{sun}

This team proposes a lightweight ensemble framework that combines a WaveMamba-based all-in-one restoration stream with NAFNet-based expert refinement. The main stream employs WaveMamba as the unified restoration backbone and introduces category-aware routing for blur and snow images, which are processed by dedicated NAFNet experts. The auxiliary stream uses an all-category NAFNet model trained to refine images affected by different degradation types. The predictions from the two streams are combined through linear image blending, where the WaveMamba and expert output contributes 95 percent and the auxiliary NAFNet output contributes 5 percent. This design retains the general restoration capability of WaveMamba while using NAFNet to provide complementary local correction and detail enhancement. The blending weight is selected according to feedback obtained from the Codabench testing phase.

\noindent \textbf{Training details.}
The restoration models are developed using the official LoViF 2026 training data together with publicly available image restoration datasets and pretrained restoration resources. The main restoration stream is based on WaveMamba, while NAFNet and NAFNetLocal are used for the auxiliary and degradation-specific expert branches. The submitted factsheet does not report the exact external datasets, optimization settings, learning-rate schedule, patch size, batch size, loss functions, or total number of training iterations.

\noindent \textbf{Testing details.}
During inference, the main WaveMamba stream processes all test images, while blur and snow inputs may be routed to their corresponding NAFNet experts. An additional all-category NAFNet model generates the auxiliary restoration result. The final prediction is obtained by blending 95 percent of the main-stream output with 5 percent of the auxiliary NAFNet output. The restored images are saved as RGB JPEG files with a quality value of 96, disabled chroma subsampling, and optimized encoding. The reported runtime is approximately 0.20 seconds per image on a GPU. Their final submission achieves an overall testing score of 29.86.

\subsection{Ipara}

This team proposes SupMoE, a supervised mixture-of-experts framework for real-world all-in-one image restoration. The network contains a fast encoder, three Transformer-based refinement blocks, a supervised mixture-of-experts module, and a PixelShuffle decoder. The encoder employs 20 gated one-dimensional convolutional blocks to efficiently extract local dependencies with limited memory consumption. These blocks combine depthwise convolution with feature gating similar to SwiGLU. The subsequent Transformer modules are equipped with active degradation perception gates to refine local structures and restoration details. The central SupMoE module contains a router that predicts probabilities for blur, haze, low-light, rain, and snow. The two degradation categories with the highest probabilities are selected for soft expert routing, and their corresponding expert outputs are adaptively combined. Unlike conventional unsupervised mixture-of-experts routing, the degradation router is directly supervised using the ground-truth degradation categories during training. Finally, a PixelShuffle decoder with an upsampling factor of four reconstructs the restored image.

\begin{figure}[t]
    \centering
    \includegraphics[width=\linewidth]{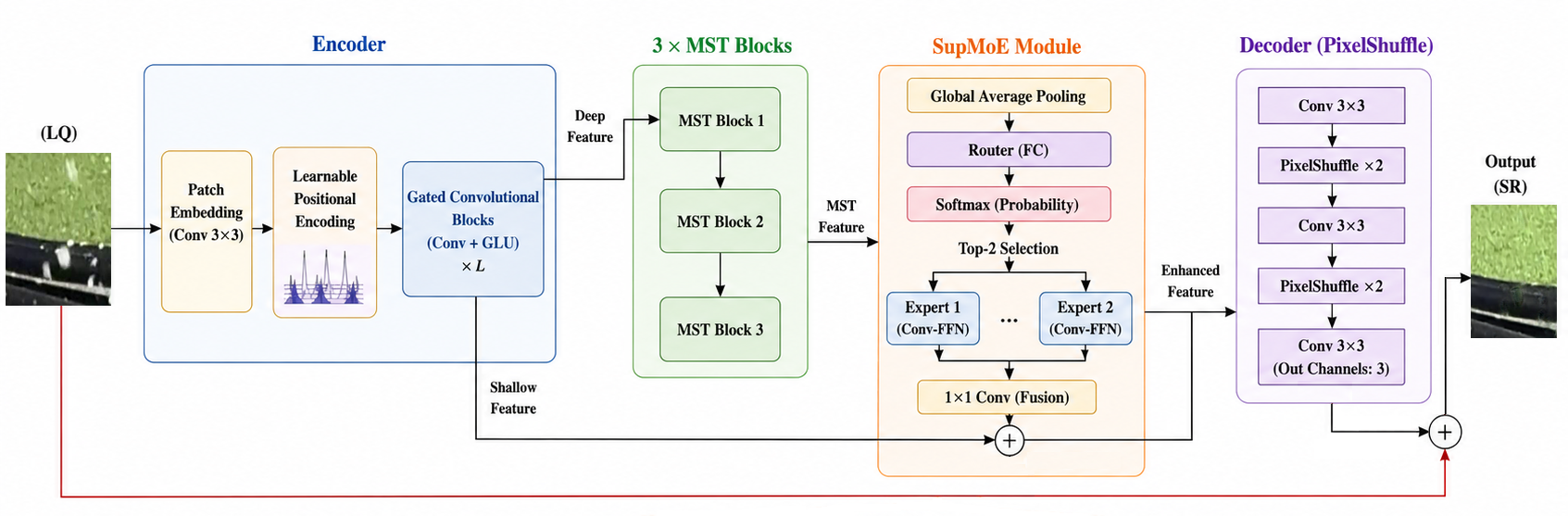}
    \caption{The overall architecture of SupMoE proposed by Team Ipara.}
    \label{fig:ipara-supmoe}
\end{figure}

\noindent \textbf{Training details.}
The model is trained using only the official LoViF training data without additional external datasets or synthetic degradation generation. The final ten images from each degradation category are held out for validation, resulting in a validation set of 50 images. Random cropping with a patch size of 128 and random horizontal and vertical flipping are used for data augmentation. Training is conducted in three phases on a single NVIDIA RTX 3090 GPU with 24 GB memory. AdamW is used throughout all phases, with gradient clipping set to 1.0 and gradient accumulation used to maintain an effective batch size of 128. In the first phase, the model is trained for 200 epochs with a constant learning rate of 0.0001. In the second phase, training continues for 390 epochs using the same learning rate so that the supervised router can learn degradation-specialized expert paths. In the final phase, the best checkpoint from the second phase is fine-tuned for 25 epochs using ReduceLROnPlateau with a reduction factor of 0.8 and a patience of 2 epochs. The weights of the router classification and auxiliary routing losses are linearly reduced by five percent per epoch and become zero after 20 epochs, allowing the final optimization to focus entirely on pixel-level restoration. During the first two phases, the objective combines L1 reconstruction loss, cross-entropy degradation classification loss, and an auxiliary routing loss with weights of 1.1, 0.008, and 0.0001, respectively.

\noindent \textbf{Testing details.}
During inference, all degraded images are restored using one SupMoE checkpoint without model ensembling, checkpoint averaging, exponential moving average weights, or output-level fusion. To process high-resolution inputs under limited GPU memory, tiled forward inference is employed with a tile size of 256 and an overlap of 20 pixels. The router predicts the degradation probabilities directly from the input features and softly activates the two most relevant restoration experts. The expert features are then fused inside the network and reconstructed by the PixelShuffle decoder. Their final submission achieves an overall testing score of 29.25.

\subsection{ustc\_pi\_lab}

This team proposes a two-stage unified restoration framework consisting of a clean DiffUIR backbone and a lightweight MoCE-Prompt residual refiner. The first stage employs DiffUIR~\cite{zheng2024diffuir} as a universal restoration backbone and initializes it from the publicly released universal checkpoint. After fine-tuning on the official LoViF training data, the milestone-310 model is selected as the clean restoration backbone. The second stage introduces a compact residual refinement network inspired by mixture-of-experts restoration designs~\cite{zheng2024diffuir}. The refiner receives a 9-channel input formed by concatenating the degraded image, the intermediate DiffUIR result, and their residual difference. A shallow convolutional stem projects this input into a feature space with a width of 32, followed by four MoCE-Prompt restoration blocks. Each block contains a shared feature-processing pathway and several lightweight expert branches with different receptive fields and processing complexities. Global pooled features are passed through a prompt and gating network to generate sample-adaptive expert weights, allowing the refiner to softly adjust its behavior for different degradations without explicit category labels. The final residual and mask heads predict a local correction and its spatial application region, and the correction is conservatively added to the DiffUIR output with a residual scale of 0.15. Both stages are packed into a single checkpoint and are applied sequentially to every input image.

\noindent \textbf{Training details.}
The model is trained using only the official LoViF paired training data, without additional external image pairs. The publicly released DiffUIR checkpoint is used for initialization, and a deterministic split with approximately 90 percent of each degradation category for training and 10 percent for validation is adopted. The DiffUIR backbone is fine-tuned using its original training objective with patches of size 256 by 256, a batch size of 8, gradient accumulation over 2 steps, and a learning rate of 0.00001. Low-light histogram equalization is disabled to maintain consistency between training and inference. After fine-tuning, the intermediate outputs of the clean backbone are cached for efficient refiner training. The residual refiner is first trained for 100,000 iterations with a learning rate of 0.0001 and is then continued to 150,000 iterations with a reduced learning rate of 0.00005. The refiner uses patches of size 256 by 256 and a batch size of 8. Its objective combines Charbonnier reconstruction loss, gradient and Laplacian structure losses, luminance and exposure losses, degradation-region-weighted reconstruction loss, snow-cue supervision, backbone-output preservation loss, and residual-magnitude regularization. The continuation stage increases the emphasis on haze and snow residual regions while maintaining conservative corrections for blur, rain, text, and structural details.

\noindent \textbf{Testing details.}
During inference, a single packed checkpoint containing both the DiffUIR backbone and the MoCE-Prompt refiner is loaded. Each test image is first processed by the clean DiffUIR backbone and is subsequently refined through the masked residual correction module. The same two-stage pipeline is used for blur, low-light, haze, rain, and snow images without degradation labels, file-index routing, task-specific checkpoints, test-time augmentation, or model ensembling. The restored images retain the original spatial resolution and file names and are saved as RGB JPEG files with a quality value of 96, disabled chroma subsampling, and optimized encoding. Their final submission achieves an overall score of 28.47, with a PSNR of 22.65, an SSIM of 0.77, and an LPIPS value of 0.38.


\section*{Teams and Affiliations}

\subsection*{Organizers}
\noindent  \textit{\textbf{Title:}} The Second LoViF 2026 Challenge on Real-World All-in-One Image Restoration

\noindent  \textit{\textbf{Members:}}
Xiang Chen (\textcolor{magenta}{chenxiang@njust.edu.cn}), Hao Li, Jiangxin Dong, and Jinshan Pan

\noindent  \textit{\textbf{Affiliations:}}
Nanjing University of Science and Technology

\subsection*{Re:Pixel}

\noindent  \textit{\textbf{Members:}}
\noindent   Hongbo Ding (\textcolor{magenta}{dinghongbo324@gmail.com})

\noindent  \textit{\textbf{Affiliations:}}
\noindent Hunan University Of Technology and Business

\subsection*{REDnoteMediaLab}

\noindent  \textit{\textbf{Members:}}
\noindent   Junpeng Jiang (\textcolor{magenta}{jjunpeng1122@outlook.com}), Xingyu Qiu, Yilian Zhong, Yuxiang Chen, Shibo Yin, Zixuan Huang, Yushun Fang, Xilei Zhu, Yahui Wang, and Chen Lu 

\noindent  \textit{\textbf{Affiliations:}}
\noindent Xiaohongshu Inc

\subsection*{LucidWorld}

\noindent  \textit{\textbf{Members:}}
\noindent   Xiaodong Zhou (\textcolor{magenta}{zhouxdshpc@ 163.com}), Qingyue Cao, and Changwei Gong

\noindent  \textit{\textbf{Affiliations:}}
\noindent People’s Public Security University of China, Shanghai Police College

\subsection*{Pheonix}

\noindent  \textit{\textbf{Members:}}
\noindent   Jingyun Liu (\textcolor{magenta}{zhouxdshpc@ 163.com}), Xingchen Yi, Hansen Shi, Ruiyi Liu, Jirui Xie, Tao Liu, and Wenzhuo Ma 

\noindent  \textit{\textbf{Affiliations:}}
\noindent Wuhan University

\subsection*{SeeIR}

\noindent  \textit{\textbf{Members:}}
\noindent   Hongzhen Li (\textcolor{magenta}{zhouxdshpc@ 163.com}), Yongyong Chen, Zheng Zhou, Jingyong Su, Jie Liu, and Haijin Zeng

\noindent  \textit{\textbf{Affiliations:}}
\noindent Harbin Institute of Technology, Guangzhou University

\subsection*{MiVideo}

\noindent  \textit{\textbf{Members:}}
\noindent   Cheng Li, Peishuai Zha, Ziyi Wang, Jian Tang, Yan Chen, Long Bao, and Heng Sun

\subsection*{MiAlgo\_LM}

\noindent  \textit{\textbf{Members:}}
\noindent   Jiyuan Zhang, Shuai Liu, Wei Ding, Chengjun Guo, Yibin Huang, Xiaotao Wang, Dongqing Zou, and Lei Lei

\subsection*{Pamy}

\noindent  \textit{\textbf{Members:}}
\noindent   Xiaofeng Wang, Xiao Liu, Yulin Wu, Yuhan Zhao, Shurui Peng, and Chao Ren

\noindent  \textit{\textbf{Affiliations:}}
\noindent Sichuan University

\subsection*{jason0411202}

\noindent  \textit{\textbf{Members:}}
\noindent   Yu-Kai Wang (\textcolor{magenta}{yokaichat001@gmail.com})

\noindent  \textit{\textbf{Affiliations:}}
\noindent National Yang Ming Chiao Tung University

\subsection*{NIT-Oita}

\noindent  \textit{\textbf{Members:}}
\noindent   Kosuke Shigematsu (\textcolor{magenta}{k-shigematsu@oita-ct.ac.jp}), and Asuka Shin

\noindent  \textit{\textbf{Affiliations:}}
\noindent National Institute of Technology

\subsection*{ACVLAB}

\noindent  \textit{\textbf{Members:}}
\noindent   Rong-Lin Jian (\textcolor{magenta}{ron0410530.ai14@nycu.edu.tw}), Cheng-Jun Kang, and Jin-Hui Jiang

\noindent  \textit{\textbf{Affiliations:}}
\noindent National Yang Ming Chiao Tung University, National Cheng Kung University

\subsection*{MoEFlow}

\noindent  \textit{\textbf{Members:}}
\noindent   Jialin Zhou (\textcolor{magenta}{zhoujialin2001@gmail.com}), and Kuo Yuan

\noindent  \textit{\textbf{Affiliations:}}
\noindent Harbin Institute of Technology

\subsection*{Zzz}

\noindent  \textit{\textbf{Members:}}
\noindent   Songyu Zhang (\textcolor{magenta}{zhangsy990113@gmail.com})

\noindent  \textit{\textbf{Affiliations:}}
\noindent Jilin University

\subsection*{BaseLess}

\noindent  \textit{\textbf{Members:}}
\noindent   E B Benson (\textcolor{magenta}{benson.elamthuruthy@a.riken.jp}), Ashfaq Hussain, and Pruthvikanth AC

\noindent  \textit{\textbf{Affiliations:}}
\noindent RIKEN BioResource Centre, Cochin University of Science and Technology

\subsection*{O4A}

\noindent  \textit{\textbf{Members:}}
\noindent   Qirui Chen (\textcolor{magenta}{qiruichen92@ gmail.com}), Jinyuan Chen, and Jun Zhang 

\noindent  \textit{\textbf{Affiliations:}}
\noindent Zhejiang University, The Hong Kong University of Science and Technology

\subsection*{GKD\_IR}

\noindent  \textit{\textbf{Members:}}
\noindent   Xu Zhang (\textcolor{magenta}{zhangx0802@whu.edu.cn}), Xuhui Cao, Jiaqi Ma, Laibin Chang, Yuchun Miao, Yichu Xu, Yuanzhi Yao, Shi Chen, Yuning Cui, Huan Zhang, and Lefei Zhang

\noindent  \textit{\textbf{Affiliations:}}
\noindent Wuhan University

\subsection*{IKLab}

\noindent  \textit{\textbf{Members:}}
\noindent   Saeed Ahmad (\textcolor{magenta}{saeedahmad@iklab.ai}), Ik Hyun Lee, Jun Young Park, and Ji Hwan Yoon

\noindent  \textit{\textbf{Affiliations:}}
\noindent IK Lab Inc.

\subsection*{sun}

\noindent  \textit{\textbf{Members:}}
\noindent   Shangquan Sun (\textcolor{magenta}{shangquan.sun@ntu.edu.sg})

\noindent  \textit{\textbf{Affiliations:}}
\noindent Nanyang Technological University

\subsection*{Ipara}

\noindent  \textit{\textbf{Members:}}
\noindent   Behrooz Nobahar-Moghanlou (\textcolor{magenta}{nobahar.behrooz@gmail.com}), Majid Edalatjou, and Karim Shahi-Niyar

\noindent  \textit{\textbf{Affiliations:}}
\noindent Idea Pardazan Ipara Co.

\subsection*{ustc\_pi\_lab}

\noindent  \textit{\textbf{Members:}}
\noindent   Ruibo Zhang (\textcolor{magenta}{nobahar.behrooz@gmail.com}), Dexiang Hong, Xinyan Liu, Shengeng Tang, and Weidong Chen

\noindent  \textit{\textbf{Affiliations:}}
\noindent University of Science and Technology of China, Harbin Institute of Technology, Hefei University of Technology

%
%
\bibliographystyle{splncs04}
\bibliography{main}

@String(CVPR  = {IEEE Conf. Comput. Vis. Pattern Recog.})

@String(ECCV  = {Eur. Conf. Comput. Vis.})

@String(NeurIPS = {Adv. Neural Inform. Process. Syst.})

@String(ICLR  = {Int. Conf. Learn. Represent.})

@String(CVPRW = {IEEE Conf. Comput. Vis. Pattern Recog. Worksh.})

@String(AAAI  = {AAAI})

@String(ACMMM = {ACM Int. Conf. Multimedia})

@String(CVPR  = {CVPR})

@String(ECCV  = {ECCV})

@String(NeurIPS = {NeurIPS})

@String(ICLR  = {ICLR})

@String(CVPRW = {CVPRW})

@String(ACMMM = {ACM MM})

@article{FFC,
  author       = {Zechen Liu and Feiyang Zhang and Wei Song and Xiang Li and Wei Wei},
  title        = {Fast Fourier Correlation is a Highly Efficient and Accurate Feature Attribution Algorithm from the Perspective of Control Theory and Game Theory},
  journal      = {CoRR},
  volume       = {abs/2504.02016},
  year         = {2025},
  eprint       = {2504.02016},
  archivePrefix= {arXiv},
  primaryClass = {cs.LG}
}

@misc{GGT100k,
  author       = {VCLab, The Hong Kong Polytechnic University},
  title        = {GGT-100K Dataset},
  howpublished = {\url{https://huggingface.co/datasets/VCLab-PolyU/GGT-100K}},
  year         = {2025},
  note         = {Hugging Face Dataset}
}

@inproceedings{zhang2018lpips,
  author    = {Richard Zhang and Phillip Isola and Alexei A. Efros and Eli Shechtman and Oliver Wang},
  title     = {The Unreasonable Effectiveness of Deep Features as a Perceptual Metric},
  booktitle = CVPR,
  pages     = {586--595},
  year      = {2018}
}

@inproceedings{chen2022nafnet,
  title={Simple Baselines for Image Restoration},
  author={Chen, Liangyu and Chu, Xiaojie and Zhang, Xiangyu and Sun, Jian},
  booktitle={Proceedings of the European Conference on Computer Vision},
  year={2022}
}

@inproceedings{nah2017deep,
  title={Deep Multi-scale Convolutional Neural Network for Dynamic Scene Deblurring},
  author={Nah, Seungjun and Kim, Tae Hyun and Lee, Kyoung Mu},
  booktitle={Proceedings of the IEEE Conference on Computer Vision and Pattern Recognition},
  pages={3883--3891},
  year={2017}
}

@inproceedings{shen2019human,
  title={Human-Aware Motion Deblurring},
  author={Shen, Ziyi and Wang, Wenguan and Lu, Xiankai and Shen, Jianbing and Ling, Haibin and Xu, Tingfa and Shao, Ling},
  booktitle={Proceedings of the IEEE/CVF International Conference on Computer Vision},
  pages={5572--5581},
  year={2019}
}

@article{cai2018learning,
  title={Learning a Deep Single Image Contrast Enhancer from Multi-exposure Images},
  author={Cai, Jianrui and Gu, Shuhang and Zhang, Lei},
  journal={IEEE Transactions on Image Processing},
  volume={27},
  number={4},
  pages={2049--2062},
  year={2018}
}

@inproceedings{skff2020,
author = {Syed Waqas Zamir and Aditya Arora and Salman Khan and Munawar Hayat and Fahad Shahbaz Khan and Ming-Hsuan Yang and Ling Shao},
title = {Learning Enriched Features for Real Image Restoration and Enhancement},
booktitle = ECCV,
year = 2020
}

@inproceedings{weatherbench,
author = {Qiyuan Guan and Qianfeng Yang and Xiang Chen and Tianyu Song and Guiyue Jin and Jiyu Jin},
title = {WeatherBench: A Real-World Benchmark Dataset for All-in-One Adverse Weather Image Restoration},
booktitle = ACMMM,
year = 2025
}

@inproceedings{mambair2024,
author = {Yong Guo and Jian Chen and Zihao Wang and Xiaojie Jin and Jiashi Feng},
title = {MambaIRv2: Attentive State Space Models for Efficient Image Restoration},
booktitle = ECCV,
year = 2024
}

@inproceedings{chen2026foundir,
  title={Foundir-v2: Optimizing pre-training data mixtures for image restoration foundation model},
  author={Chen, Xiang and Pan, Jinshan and Dong, Jiangxin and Yang, Jian and Tang, Jinhui},
  booktitle={Proceedings of the IEEE/CVF Conference on Computer Vision and Pattern Recognition},
  pages={8471--8480},
  year={2026}
}

@inproceedings{guan2025weatherbench,
  title={Weatherbench: A real-world benchmark dataset for all-in-one adverse weather image restoration},
  author={Guan, Qiyuan and Yang, Qianfeng and Chen, Xiang and Song, Tianyu and Jin, Guiyue and Jin, Jiyu},
  booktitle={Proceedings of the 33rd ACM international conference on multimedia},
  pages={12607--12613},
  year={2025}
}

@article{datprl_ir,
  author  = {Guanglu Dong and Chunlei Li and Chao Ren and Jingliang Hu and Yilei Shi and Xiao Xiang Zhu and Lichao Mou},
  title   = {Learning Domain-Aware Task Prompt Representations for Multi-Domain All-in-One Image Restoration},
  journal = {arXiv preprint arXiv:2603.01725},
  year    = {2026}
}

@inproceedings{mobilenetv3,
  author    = {Andrew Howard and Mark Sandler and Grace Chu and Liang-Chieh Chen and Bo Chen and Mingxing Tan and Weijun Wang and Yukun Zhu and Ruoming Pang and Vijay Vasudevan and Quoc V. Le and Hartwig Adam},
  title     = {Searching for MobileNetV3},
  booktitle = {Proceedings of the IEEE/CVF International Conference on Computer Vision},
  pages     = {1314--1324},
  year      = {2019}
}

@article{cai2026hidream,
  title={Hidream-o1-image: A natively unified image generative foundation model with pixel-level unified transformer},
  author={Cai, Qi and Chen, Jingwen and Gao, Chengmin and Gong, Zijian and Li, Yehao and Pan, Yingwei and Peng, Yi and Qiu, Zhaofan and Yu, Kai and Zhang, Yiheng and others},
  journal={arXiv preprint arXiv:2605.11061},
  year={2026}
}

@inproceedings{zhang2019root,
  title={Root mean square layer normalization},
  author={Zhang, Biao and Sennrich, Rico},
  booktitle={Advances in neural information processing systems},
  volume={32},
  year={2019}
}

@article{shazeer2020glu,
  title={Glu variants improve transformer},
  author={Shazeer, Noam},
  journal={arXiv preprint arXiv:2002.05202},
  year={2020}
}

@inproceedings{ainslie2023gqa,
  title={Gqa: Training generalized multi-query transformer models from multi-head checkpoints},
  author={Ainslie, Joshua and Lee-Thorp, James and De Jong, Michiel and Zemlyanskiy, Yury and Lebr{\'o}n, Federico and Sanghai, Sumit},
  booktitle={Proceedings of the 2023 Conference on Empirical Methods in Natural Language Processing},
  pages={4895--4901},
  year={2023}
}

@article{su2024roformer,
  title={Roformer: Enhanced transformer with rotary position embedding},
  author={Su, Jianlin and Ahmed, Murtadha and Lu, Yu and Pan, Shengfeng and Bo, Wen and Liu, Yunfeng},
  journal={Neurocomputing},
  volume={568},
  pages={127063},
  year={2024},
  publisher={Elsevier}
}

@inproceedings{caron2021emerging,
  title={Emerging properties in self-supervised vision transformers},
  author={Caron, Mathilde and Touvron, Hugo and Misra, Ishan and J{\'e}gou, Herv{\'e} and Mairal, Julien and Bojanowski, Piotr and Joulin, Armand},
  booktitle={Proceedings of the IEEE/CVF international conference on computer vision},
  pages={9650--9660},
  year={2021}
}

@INPROCEEDINGS{Zamir2022Restormer,
  author={Zamir, Syed Waqas and Arora, Aditya and Khan, Salman and Hayat, Munawar and Khan, Fahad Shahbaz and Yang, Ming–Hsuan},
  booktitle={Proceedings of the IEEE/CVF Conference on Computer Vision and Pattern Recognition}, 
  title={Restormer: Efficient Transformer for High-Resolution Image Restoration}, 
  year={2022},
  pages={5718-5729}
}

@article{Pan2026XRestormerPlusPlus,
  author = {Pan, Youwei and Cao, Leilei and Zhu, Yingfang and Zhu, Fengjie},
  title = {X-Restormer++: 1st Place Solution for the UG2+ CVPR 2026 All-Weather Restoration Challenge},
  journal = {arXiv preprint arXiv:2605.13258},
  year = {2026}
}

@article{WANG2026104380,
  author = {Wang, Xiaofeng and Liu, Xiao and Yang, Yutong and Wang, Zhengyong and He, Xiaohai and Chen, Honggang and Li, Yi and Wang, Pingyu},
  title = {DSRIR: Dynamic Spatial Refinement Learning for Progressive All-in-One Image Restoration},
  journal = {Information Processing \& Management},
  volume = {63},
  number = {2, Part B},
  pages = {104380},
  year = {2026}
}

@inproceedings{Li2024FoundIR,
  author = {Li, Hao and Chen, Xiang and Dong, Jiangxin and Tang, Jinhui and Pan, Jinshan},
  title = {FoundIR: Unleashing Million-Scale Training Data to Advance Foundation Models for Image Restoration},
  booktitle = {Proceedings of the IEEE/CVF International Conference on Computer Vision},
  pages = {12626--12636},
  year = {2025}
}

@inproceedings{cai2023retinexformer,
  title={Retinexformer: One-stage retinex-based transformer for low-light image enhancement},
  author={Cai, Yuanhao and Bian, Hao and Lin, Jing and Wang, Haoqian and Timofte, Radu and Zhang, Yulun},
  booktitle={Proceedings of the IEEE/CVF international conference on computer vision},
  pages={12504--12513},
  year={2023}
}

@article{chen2023comparative,
  title={A Comparative Study of Image Restoration Networks for General Backbone Network Design}, 
  author={Chen, Xiangyu and Li, Zheyuan and Pu, Yuandong and Liu, Yihao and Zhou, Jiantao and Qiao, Yu and Dong, Chao},
  journal={arXiv preprint arXiv:2310.11881},
  year={2023}
}

@misc{simeoni2025dinov3,
  title={{DINOv3}},
  author={Sim{\'e}oni, Oriane and Vo, Huy V. and Seitzer, Maximilian and Baldassarre, Federico and Oquab, Maxime and Jose, Cijo and Khalidov, Vasil and Szafraniec, Marc and Yi, Seungeun and Ramamonjisoa, Micha{\"e}l and Massa, Francisco and Haziza, Daniel and Wehrstedt, Luca and Wang, Jianyuan and Darcet, Timoth{\'e}e and Moutakanni, Th{\'e}o and Sentana, Leonel and Roberts, Claire and Vedaldi, Andrea and Tolan, Jamie and Brandt, John and Couprie, Camille and Mairal, Julien and J{\'e}gou, Herv{\'e} and Labatut, Patrick and Bojanowski, Piotr},
  year={2025},
  eprint={2508.10104},
  archivePrefix={arXiv},
  primaryClass={cs.CV},
  url={https://arxiv.org/abs/2508.10104},
}

@inproceedings{daclip,
  title={Controlling vision-language models for multi-task image restoration},
  author={Luo, Ziwei and Gustafsson, Fredrik K and Zhao, Zheng and Sj{\"o}lund, Jens and Sch{\"o}n, Thomas},
  booktitle={ICLR},
  volume={2024},
  pages={16226--16246},
  year={2024}
}

@inproceedings{zheng2024diffuir,
  author    = {Dian Zheng and Xiao-Ming Wu and Shuzhou Yang and Jian Zhang and Jian-Fang Hu and Wei-Shi Zheng},
  title     = {Selective Hourglass Mapping for Universal Image Restoration Based on Diffusion Model},
  booktitle = {Proceedings of the IEEE/CVF Conference on Computer Vision and Pattern Recognition},
  pages     = {25445--25455},
  year      = {2024}
}

@inproceedings{mirage,
  author    = {Bin Ren and Yawei Li and Xu Zheng and Yuqian Fu and
               Danda Pani Paudel and Hong Liu and Ming-Hsuan Yang and
               Luc Van Gool and Nicu Sebe},
  title     = {Efficient Degradation-Agnostic Image Restoration via
               Channel-Wise Functional Decomposition and Manifold Regularization},
  booktitle = {International Conference on Learning Representations},
  year      = {2026}
}

@inproceedings{convnext,
  author    = {Zhuang Liu and Hanzi Mao and Chao-Yuan Wu and
               Christoph Feichtenhofer and Trevor Darrell and Saining Xie},
  title     = {A ConvNet for the 2020s},
  booktitle = {Proceedings of the IEEE/CVF Conference on Computer Vision
               and Pattern Recognition},
  pages     = {11976--11986},
  year      = {2022}
}

@inproceedings{chen2026lovif,
  title={LoViF 2026 challenge on real-world all-in-one image restoration: Methods and results},
  author={Chen, Xiang and Li, Hao and Dong, Jiangxin and Pan, Jinshan and Li, Xin and He, Xin and Chen, Naiwei and Li, Shengyuan and Liu, Fengning and Lv, Haoyi and others},
  booktitle={Proceedings of the IEEE/CVF Conference on Computer Vision and Pattern Recognition},
  pages={4927--4935},
  year={2026}
}

@inproceedings{li2024promptcir,
  author    = {Li, Bingchen and Li, Xin and Lu, Yiting and Feng, Ruoyu and Guo, Mengxi and Zhao, Shijie and Zhang, Li and Chen, Zhibo},
  title     = {{PromptCIR}: Blind Compressed Image Restoration with Prompt Learning},
  booktitle = CVPRW,
  year      = {2024}
}

@inproceedings{potlapalli2023promptir,
  author    = {Potlapalli, Vaishnav and Zamir, Syed Waqas and Khan, Salman and Khan, Fahad Shahbaz},
  title     = {{PromptIR}: Prompting for All-in-One Blind Image Restoration},
  booktitle = NeurIPS,
  year      = {2023}
}

@article{gu2023mamba,
  author  = {Gu, Albert and Dao, Tri},
  title   = {Mamba: Linear-Time Sequence Modeling with Selective State Spaces},
  journal = {arXiv:2312.00752},
  year    = {2023}
}

@article{liu2025dpmambair,
  author  = {Liu, Zhanwen and others},
  title   = {{DPMambaIR}: All-in-One Image Restoration via Degradation-Aware Prompt State Space Model},
  journal = {arXiv:2504.17732},
  year    = {2025}
}

@inproceedings{kong2025evssm,
  author    = {Kong, Lingshun and Dong, Jiangxin and Tang, Jinhui and Yang, Ming-Hsuan and Pan, Jinshan},
  title     = {Efficient Visual State Space Model for Image Deblurring},
  booktitle = CVPR,
  year      = {2025}
}

@inproceedings{feijoo2025darkir,
  author    = {Feijoo, Daniel and Benito, Juan C. and Garcia, Alvaro and Conde, Marcos V.},
  title     = {{DarkIR}: Robust Low-Light Image Restoration},
  booktitle = CVPR,
  pages     = {10879--10889},
  year      = {2025}
}

@inproceedings{yan2025hvi,
  author    = {Yan, Qingsen and others},
  title     = {{HVI}: A New Color Space for Low-Light Image Enhancement},
  booktitle = CVPR,
  year      = {2025}
}

@article{jiang2025survey,
  title={A survey on all-in-one image restoration: Taxonomy, evaluation and future trends},
  author={Jiang, Junjun and Zuo, Zengyuan and Wu, Gang and Jiang, Kui and Liu, Xianming},
  journal={IEEE TPAMI},
  year={2025},
  publisher={IEEE}
}

@inproceedings{Airnet,
  title={All-in-one image restoration for unknown corruption},
  author={Li, Boyun and Liu, Xiao and Hu, Peng and Wu, Zhongqin and Lv, Jiancheng and Peng, Xi},
  booktitle={CVPR},
  year={2022}
}

@inproceedings{chen2023learning,
  title={Learning a sparse transformer network for effective image deraining},
  author={Chen, Xiang and Li, Hao and Li, Mingqiang and Pan, Jinshan},
  booktitle={CVPR},
  year={2023}
}

@inproceedings{kong2023efficient,
  title={Efficient frequency domain-based transformers for high-quality image deblurring},
  author={Kong, Lingshun and Dong, Jiangxin and Ge, Jianjun and Li, Mingqiang and Pan, Jinshan},
  booktitle={CVPR},
  year={2023}
}

@inproceedings{li2025ntire,
  title={NTIRE 2025 challenge on day and night raindrop removal for dual-focused images: Methods and results},
  author={Li, Xin and Jin, Yeying and Jin, Xin and Wu, Zongwei and Li, Bingchen and Wang, Yufei and Yang, Wenhan and Li, Yu and Chen, Zhibo and Wen, Bihan and others},
  booktitle={CVPR Workshop},
  year={2025}
}

@article{cui2026bio,
  title={Bio-inspired image restoration},
  author={Cui, Yuning and Ren, Wenqi and Knoll, Alois},
  journal={Advances in Neural Information Processing Systems},
  volume={38},
  pages={80452--80481},
  year={2026}
}

@inproceedings{cui2025adair,
  title={Adair: Adaptive all-in-one image restoration via frequency mining and modulation},
  author={Cui, Yuning and Zamir, Syed Waqas and Khan, Salman and Knoll, Alois and Shah, Mubarak and Khan, Fahad},
  booktitle={International Conference on Learning Representations},
  volume={2025},
  pages={101306--101327},
  year={2025}
}

@inproceedings{jin2025smokebench,
  title={SmokeBench: A Real-World Dataset for Surveillance Image Desmoking in Early-Stage Fire Scenes},
  author={Jin, Wenzhuo and Yang, Qianfeng and Wu, Xianhao and Chen, Hongming and Li, Pengpeng and Chen, Xiang},
  booktitle={Proceedings of the 33rd ACM International Conference on Multimedia},
  pages={12722--12728},
  year={2025}
}

@article{guan2026rethinking,
  title={Rethinking nighttime image deraining via learnable color space transformation},
  author={Guan, Qiyuan and Chen, Xiang and Jin, Guiyue and Jin, Jiyu and Fan, Shumin and Song, Tianyu and Pan, Jinshan},
  journal={Advances in Neural Information Processing Systems},
  volume={38},
  pages={3189--3225},
  year={2026}
}

@article{guan2026harmonizing,
  title={Harmonizing Light and Darkness: Nighttime Image Deraining Using Color Space Transformation},
  author={Guan, Qiyuan and Fan, Shumin and Jin, Jiyu and Jin, Guiyue and Song, Tianyu and Li, Pengpeng and Chen, Xiang and Jiang, Kui},
  journal={IEEE Transactions on Multimedia},
  year={2026},
  publisher={IEEE}
}

@inproceedings{yang2026rethinking,
  title={Rethinking rainy 3D scene reconstruction via perspective transforming and brightness tuning},
  author={Yang, Qianfeng and Chen, Xiang and Li, Pengpeng and Guan, Qiyuan and Jin, Guiyue and Jin, Jiyu},
  booktitle={Proceedings of the AAAI Conference on Artificial Intelligence},
  volume={40},
  number={14},
  pages={11658--11666},
  year={2026}
}

@inproceedings{yang2026unirain,
  title={UniRain: Unified Image Deraining with RAG-based Dataset Distillation and Multi-objective Reweighted Optimization},
  author={Yang, Qianfeng and Guan, Qiyuan and Chen, Xiang and Jin, Jiyu and Jin, Guiyue and Dong, Jiangxin},
  booktitle={Proceedings of the IEEE/CVF Conference on Computer Vision and Pattern Recognition},
  pages={12428--12437},
  year={2026}
}

@article{yang2026textual,
  title={Textual--visual interaction for enhanced single image deraining using adapter-tuned VLMs},
  author={Yang, Qianfeng and Li, Pengpeng and Jin, Jiyu and Jin, Guiyue and Song, Tianyu and Fan, Shumin and Hou, Hao},
  journal={The Visual Computer},
  volume={42},
  number={4},
  pages={193},
  year={2026},
  publisher={Springer}
}
\end{document}